%% file: main.tex
\documentclass{article} % For LaTeX2e
\usepackage{arxiv,times}
\input{math_commands.tex}

\usepackage{wrapfig}
\usepackage{amsmath,amssymb,amsfonts}
\usepackage{algorithmic}
\usepackage{graphicx}
\usepackage[numbers]{natbib}
\usepackage{textcomp}
\usepackage{xcolor}
\def\BibTeX{{\rm B\kern-.05em{\sc i\kern-.025em b}\kern-.08em
    T\kern-.1667em\lower.7ex\hbox{E}\kern-.125emX}}

\usepackage[utf8]{inputenc} % allow utf-8 input
\usepackage[T1]{fontenc}    % use 8-bit T1 fonts
\usepackage{hyperref} 
\usepackage{booktabs}       
\usepackage{array}
\usepackage{algorithm}

\usepackage[caption=false]{subfig}
\usepackage{caption}
\newcommand{\method}{CryptoMamba}

\begin{document}

\title{\method{}: Leveraging State Space Models for Accurate Bitcoin Price Prediction\\
}

\author{Mohammad Shahab Sepehri$^{\dagger}$ \quad
Asal Mehradfar$^{\dagger}$ \quad
Mahdi Soltanolkotabi \quad
Salman Avestimehr \\
Department of Electrical and Computer Engineering, University of Southern California \\
\texttt{\{sepehri, mehradfa, soltanol, avestime\}@usc.edu}
}

\renewcommand{\thefootnote}{}
% \makeatletter\Hy@raisedlink{\relax}\makeatother
\footnotetext{Published in IEEE International Conference on Blockchain and Cryptocurrency (ICBC) 2025.}
\def\thefootnote{\textdagger}\footnotetext[0]{These authors contributed equally to this work.}\def\thefootnote{\arabic{footnote}}

% \author{
%   Mohammad Shahab Sepehri$^{\dagger}$ \\
%   Department of Electrical and Computer Engineering \\
%   University of Southern California \\
%   \texttt{sepehri@usc.edu} \\
%   \And
%   Asal Mehradfar$^{\dagger}$ \\
%   Department of Electrical and Computer Engineering \\
%   University of Southern California \\
%   \texttt{mehradfa@usc.edu} \\
%   \And
%   Mahdi Soltanolkotabi \\
%   Department of Electrical and Computer Engineering \\
%   University of Southern California \\
%   \texttt{soltanol@usc.edu} \\
%   \And
%   Salman Avestimehr \\
%   Department of Electrical and Computer Engineering \\
%   University of Southern California \\
%   \texttt{avestime@usc.edu} \\
% }

\maketitle

\input{sections/0_abstract}
% \begin{abstract}
% This document is a model and instructions for \LaTeX.
% This and the IEEEtran.cls file define the components of your paper [title, text, heads, etc.]. *CRITICAL: Do Not Use Symbols, Special Characters, Footnotes, 
% or Math in Paper Title or Abstract.
% \end{abstract}

\input{sections/1_introduction}
\input{sections/2_related}
\input{sections/3_background}
\input{sections/4_method}
\input{sections/5_experiments}

\input{sections/6_application}
\input{sections/7_conclusion}
% \newpage
\bibliography{references}
\bibliographystyle{unsrt}

\newpage
\appendix
\input{sections/8_appendix}

\end{document}

%% file: math_commands.tex
\usepackage{amsmath,amsfonts,bm}

\def\eqref#1{equation~\ref{#1}}
\def\1{\bm{1}}

\DeclareMathAlphabet{\mathsfit}{\encodingdefault}{\sfdefault}{m}{sl}
\SetMathAlphabet{\mathsfit}{bold}{\encodingdefault}{\sfdefault}{bx}{n}

%% file: sections/0_abstract.tex
\begin{abstract}

Predicting Bitcoin price remains a challenging problem due to the high volatility and complex non-linear dynamics of cryptocurrency markets. Traditional time-series models, such as ARIMA and GARCH, and recurrent neural networks, like LSTMs, have been widely applied to this task but struggle to capture the regime shifts and long-range dependencies inherent in the data. In this work, we propose \method{}, a novel Mamba-based State Space Model (SSM) architecture designed to effectively capture long-range dependencies in financial time-series data. Our experiments show that \method{} not only provides more accurate predictions but also offers enhanced generalizability across different market conditions, surpassing the limitations of previous models. 
Coupled with trading algorithms for real-world scenarios, \method{} demonstrates its practical utility by translating accurate forecasts into financial outcomes. Our findings signal a huge advantage for SSMs in stock and cryptocurrency price forecasting tasks. The codebase is available in the following link: \url{https://github.com/MShahabSepehri/CryptoMamba}.

\end{abstract}

%% file: sections/1_introduction.tex
\section{Introduction}\label{sec:intro}

Predicting Bitcoin~\citep{nakamoto2008bitcoin} prices is a critical problem due to the high volatility of the cryptocurrency market \citep{jang2017empirical}. With the growing influence of cryptocurrencies, the demand for reliable prediction models continues to rise, especially for those seeking to capitalize on market opportunities or safeguard against losses. A successful solution to this problem would benefit traders, institutions, and regulators by offering deeper insights into market behavior, enhancing decision-making, and improving the stability of cryptocurrency markets.

The main challenge in Bitcoin price prediction lies in capturing the complexity, nonlinearity, and long-range dependencies within the data. Bitcoin's price movements are driven by a wide range of factors, including market sentiment \citep{jagini2023sentiment}, regulatory developments \citep{aalborg2019can}, and macroeconomic trends \citep{kristoufek2013bitcoin}. These factors interact in unpredictable ways, introducing non-stationarity into the data, making accurate forecasting particularly difficult. 

Addressing these challenges requires models capable of effectively modeling temporal dependencies and adapting to dynamic market conditions. However, traditional statistical methods, such as ARIMA \citep{box2015time} and GARCH \citep{bollerslev1986generalized}, often fall short in handling complex non-linearities and sudden regime changes. In contrast, deep learning models like LSTMs (Long Short-Term Memory) \citep{lstm} and Transformers \citep{vaswani2017attention} show promise in learning complex nonlinear patterns but remain limited in scalability and generalizability.

State Space Models (SSMs) \citep{gu2021combining} provide a promising alternative by modeling time series data as a combination of latent states and observed variables. These models are well-suited for handling the temporal and stochastic characteristics of financial data. Recent advances in SSMs, such as \citep{gu2021efficiently}, demonstrate their capability to capture long-range dependencies in sequences more effectively than traditional recurrent models. Despite their success in natural language processing \citep{gu_mamba_2023, mamba2} and computer vision \citep{sepehri2024serpent}, the application of SSMs to financial time series, particularly in cryptocurrency markets, remains unexplored.

In this work, we introduce \textbf{\method{}}, which, to the best of our knowledge, is the first framework to leverage State Space Models (SSMs) to tackle the challenges of Bitcoin price prediction and among the first to apply Mamba-based models for time-series forecasting. Specifically, we:
\begin{itemize}
    \item Propose \method{}, a robust novel Mamba-based SSM architecture for capturing long-range dependencies in financial time-series data.
    \item Investigate the effects of volume as an input on the accuracy of the prediction.
    \item Define and evaluate three trading algorithms, Vanilla, Smart, and Extended Smart, to assess the real-world application of predictions.
    \item Compare \method{} against multiple baselines, demonstrating its superior performance in forecasting accuracy, financial returns, and computational efficiency.
\end{itemize}
Our work bridges the gap between advancements in SSMs and their practical financial applications, paving the way for future research in adaptive and robust market forecasting techniques.

%% file: sections/2_related.tex
\section{Related Work}\label{sec:related}

Early attempts at Bitcoin price forecasting predominantly employed classical time series models such as ARIMA (AutoRegressive Integrated Moving Average) and GARCH (Generalized Autoregressive Conditional Heteroskedasticity), widely adopted for financial time series prediction due to their ability to model linear relationships and volatility clustering, respectively. While ARIMA models, introduced by \citep{box2015time}, are commonly used in financial forecasting \citep{wirawan2019short, tian2023arima, kumar2021arima}, they assume constant variance for the data, limiting their ability to model time-varying volatility. To address this limitation, GARCH models \citep{bollerslev1986generalized} have been employed, as they excel at capturing volatility clustering, a key feature in financial markets. For instance, \citep{katsiampa2017volatility, chu2017garch} applied GARCH to model Bitcoin’s volatility, demonstrating its effectiveness for short-term forecasting. However, both ARIMA and GARCH struggle with the non-linearities and sudden regime changes that are typical of cryptocurrency markets.

To address these limitations, more recent studies have turned to machine learning techniques \citep{oreshkin2019n, lim2021temporal} for time-series prediction tasks. Among these, LSTM (Long Short-Term Memory) networks and GRU (Gated Recurrent Units) have gained prominence due to their ability to model sequential dependencies. \citep{alessandretti2018anticipating} uses LSTMs to forecast cryptocurrency prices and shows that deep learning models can outperform traditional methods in capturing complex temporal dependencies. \citep{seabe2023forecasting} compares the effectiveness of three deep learning models, LSTM, GRU, and Bi-Directional LSTM (Bi-LSTM), for predicting cryptocurrency prices. Focusing on Bitcoin, Ethereum, and Litecoin, the study finds that Bi-LSTM provides the most accurate predictions, outperforming the other models. Nonetheless, these models are often prone to overfitting and require large amounts of data to generalize well, limiting their applicability in highly volatile markets with limited data, such as Bitcoin.

In addition to recurrent architectures, Transformer models have gained traction in time-series forecasting due to their ability to model long-range dependencies via self-attention mechanisms. iTransformer~\citep{liu2023itransformer} adapts the Transformer architecture specifically for time-series data by introducing frequency-enhanced and hierarchical attention mechanisms, achieving state-of-the-art performance across several benchmarks. However, in the context of financial forecasting, particularly for volatile and non-stationary assets like Bitcoin, Transformer-based models often suffer from overfitting and lack robustness when trained on limited data. These shortcomings make them less suitable for high-volatility markets, motivating the use of more structured and efficient alternatives.

Additionally, \citep{pabucccu2023forecasting} applies various machine learning algorithms, including Support Vector Machines (SVM), Artificial Neural Networks (ANN), Naïve Bayes (NB), and Random Forest (RF), to the Bitcoin price prediction task. This study underscores the effectiveness of machine learning techniques in forecasting cryptocurrency prices and highlights the potential for improving prediction accuracy through modern modeling approaches. Similarly, \citep{amjad2017trading} utilizes machine learning classification models to predict whether Bitcoin prices will increase or decrease, providing insights into their application for real-time trading decisions.

Despite advances in machine learning and time series models, existing approaches often struggle to effectively capture the long-range dependencies and regime shifts that characterize cryptocurrency markets. One promising approach to address the stochastic nature of Bitcoin price is State Space Models (SSMs) \citep{gu2021efficiently, gu2021combining}. SSMs offer a robust framework for time series analysis, particularly well-suited for capturing long-range dependencies \citep{gu2021efficiently}, making them strong candidates for Bitcoin price prediction. Unlike traditional methods such as ARIMA and GARCH, which focus on linear relationships and short-term dependencies \citep{chu2017garch, wirawan2019short}, SSMs provide greater flexibility in modeling non-linearities and complex temporal interactions. Furthermore, compared to recurrent neural networks (e.g., LSTMs and GRUs), SSMs are computationally efficient, enabling them to better handle dependencies across longer time horizons. These advantages motivate the investigation of SSMs in the context of Bitcoin price forecasting.

Recently, Mamba \citep{gu_mamba_2023}, a novel variant of SSM, has gained attention for introducing a selectivity mechanism that adapts model parameters to the input data. Building on this foundation, \citep{wang2024mamba} proposed S-Mamba, a model specifically designed for time series forecasting. S-Mamba employs a bidirectional Mamba layer to encode inter-variate correlations and a feed-forward network to extract temporal dependencies. S-Mamba not only outperforms state-of-the-art models in accuracy but also significantly reduces computational overhead, making it particularly effective for high-dimensional time series forecasting tasks. These advancements suggest Mamba-based models as promising candidates for financial applications like Bitcoin price forecasting.

%% file: sections/3_background.tex
\section{Background}\label{sec:background}

State Space Models (SSMs) \citep{gu2021efficiently, gu2021combining} are a recent class of sequence models with roots in control theory \citep{kalman1960new}. SSMs combine the advantages of Recurrent Neural Networks (RNNs) and Convolutional Neural Networks (CNNs), making them highly effective for capturing long-range dependencies in time series data. In particular, SSMs process 1-D input sequences, where each element of the array can interact with previously scanned elements through a low-dimensional hidden state. These models are discretizations of continuous-time systems described by:
\begin{gather*}
    \dot{x} = Ax(t) + Bu(t) \\
    y(t) = Cx(t) + Du(t),
\end{gather*}
where $u \in \mathbb{R}$ is the input, $x \in \mathbb{R}^N$ is the hidden state, $y \in \mathbb{R}$ is the output, and $A \in \mathbb{R}^{N \times N}$, $B \in \mathbb{R}^{N \times 1}$, $C \in \mathbb{R}^{1 \times N}$, and $D \in \mathbb{R}$ are model parameters. In discrete time, these equations become:
\begin{gather*}
	x_k = \bar{A} x_{k-1} + \bar{B}u_k \\
	y_k = Cx_k + Du_k,
\end{gather*}
where $\bar{A}$, $\bar{B}$, and $\bar{C}$ are derived through discretization methods like zero-order hold for a given time step $\Delta$, and $D$ is often omitted. 

Traditional SSMs are linear time-invariant (LTI) systems where model dynamics do not depend on the inputs. Mamba \citep{gu_mamba_2023} added context awareness to SSMs by making $B$, $C$, and $\Delta$ input-dependent, creating a time-varying system. Moreover, Mamba leverages a hardware-aware algorithm to maintain computational efficiency, and as a result, its computational cost scales linearly with sequence length (similar to traditional SSMs). Empirical results demonstrate that Mamba achieves state-of-the-art performance in language, audio, and genomic tasks, outperforming both standard SSMs and Transformers. 
Inspired by Mamba's advancements, our approach is designing a custom Mamba-based architecture tailored for time-series forecasting.

%% file: sections/4_method.tex
\section{Methodology}\label{sec:method}

\method{} leverages Mamba-based SSMs to tackle the challenges of Bitcoin price prediction, offering a robust approach to capturing long-range dependencies in highly volatile financial data. This section provides an overview of the dataset utilized and the components of \method{}.

\subsection{Dataset}

A critical gap in Bitcoin price prediction literature is the lack of unified datasets, with studies often focusing on either early adoption periods \citep{katsiampa2017volatility} or more recent years with higher trading volumes \citep{seabe2023forecasting}. 
These inconsistencies hinder the accurate assessment of the models' generalization capabilities. To address this, we use the most recent publicly available dataset, reflecting current market trends, and evaluate \method{} against baseline models to assess its effectiveness under real-world conditions.

We use historical daily Bitcoin price data from \href{https://finance.yahoo.com/quote/BTC-USD/history/}{Yahoo Finance} covering the period from September 17, 2018, to September 17, 2024, for our experiments. The data was partitioned into distinct intervals as shown in Table \ref{table:dataset}.
For our experiments, we isolate the test and the validation sets to evaluate model performance on genuinely unseen data, enabling a more robust assessment of the model’s generalization capacity.

\begin{table}[h]
\begin{minipage}[c]{0.47\textwidth}
\centering
\caption{Dataset splits}
\label{table:dataset}
\renewcommand{\arraystretch}{1.3} 
\resizebox{0.95\textwidth}{!}{
\begin{tabular}{cc}
\toprule
\textbf{Split} & \textbf{Time Interval} \\
\midrule
Train & September 17, 2018 - September 17, 2022 \\
Validation & September 17, 2022 - September 17, 2023 \\
Test & September 17, 2023 - September 17, 2024\\
\bottomrule
\end{tabular}
}
\end{minipage}
\hfill
\begin{minipage}[c]{0.49\textwidth}
\centering
\caption{Dataset specifications}
\label{table:specifications}
\renewcommand{\arraystretch}{1.3} 
\resizebox{0.9\textwidth}{!}{
\begin{tabular}{ccc}
\toprule
\textbf{Parameter} && \textbf{Description} \\
\midrule
Open && Daily opening price of BTC-USD \\
High && Daily highest price of BTC-USD \\
Low && Daily lowest price of BTC-USD \\
Close && Daily close price of BTC-USD  \\
Volume && Daily number of Bitcoin units traded  \\
Timestamp && Date of the observation \\
\bottomrule
\end{tabular}
}
\end{minipage}
\end{table}

The data includes five main features: open, close, high, low, and volume. Since the intervals in our dataset are daily, the open and close prices represent the value of Bitcoin at the start and end of each day, which in Yahoo Finance data corresponds to the UTC time zone. The high and low prices are the maximum and minimum values recorded during each day, highlighting price volatility. Volume represents the number of Bitcoin units traded over the day, providing potential insights into market activity and sentiment that may affect price trends. Table \ref{table:specifications} summarizes different features and their descriptions.

We conducted separate analyses with and without the volume data to evaluate its impact on forecast accuracy. Volume is hypothesized to be a valuable feature as it reflects trading activity, which can signal demand and market sentiment, potentially affecting price movements. However, its effectiveness in prediction remains unknown, as most previous works do not incorporate it as an input. We test both scenarios to investigate the effect of using volume data in prediction.

\subsection{\method{} Architecture}

\begin{figure*}[!ht]
    \centering
    \includegraphics[width=0.9\textwidth]{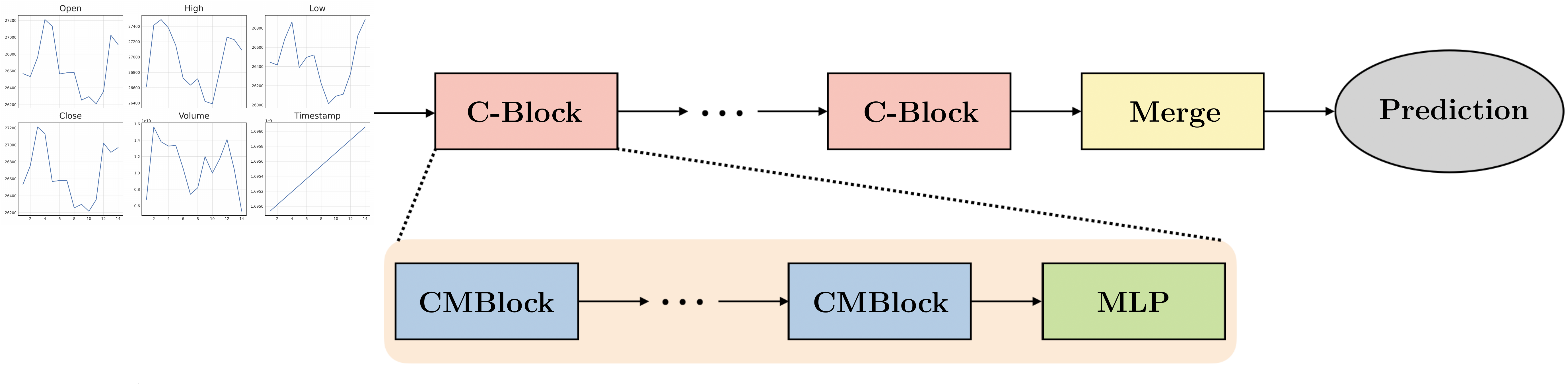}
    \caption{\method{} model consists of several C-Blocks followed by a Merge block. In each C-Block, we have several CMBlock and an MLP at the end.}
    \label{fig:model}
\end{figure*}

\method{} is a Mamba-based architecture specifically designed for financial time-series prediction, leveraging Mamba blocks to handle long-range dependencies in sequential data. The model consists of several stacked computational blocks, referred to as C-Blocks, followed by a final Merge block that generates the prediction output. The input to \method{} is the features of a fixed number of past days, and the output is the predicted closing value for the next day. The overall architecture is depicted in Figure \ref{fig:model}.

Each C-Block is composed of multiple CMBlocks and a Multi-Layer Perceptron (MLP). A CMBlock includes a normalization layer followed by a Mamba block. The output of each CMBlock feeds into subsequent CMBlocks within the same C-Block, allowing hierarchical feature extraction. The MLP at the end of each C-Block is a linear layer that adjusts the sequence length to prepare the output for the next C-Block. The outputs of all C-Blocks are aggregated by the Merge block, which is a simple linear layer to integrate the extracted features and produce the final prediction.

The hierarchical structure of \method{} enables it to progressively refine features across multiple C-Blocks, capturing both short-term and long-term temporal dependencies. Additionally, the Mamba block’s input-dependent dynamics ensure adaptability to the variability and stochastic nature of financial data.

%% file: sections/5_experiments.tex
\section{Experiments}\label{sec:experiment}

To evaluate the effectiveness of \method{} in Bitcoin price prediction, we conduct comprehensive experiments comparing its performance with several baseline models: LSTM, Bi-LSTM, GRU, iTransformer, and S-Mamba. The evaluation focuses on two aspects: prediction accuracy, measured using RMSE, MAPE, and MAE (further explained in detail in Appendix~\ref{sec:metrics}), and model efficiency, quantified by the number of parameters. Additionally, we assess the impact of incorporating trading volume as a feature by conducting experiments under two setups: with and without volume. This section presents the experimental setup, results, and analysis, providing insights into the advantages of \method{} over traditional and state-of-the-art baselines.

\subsection{Setup}

To evaluate the performance of \method{}, we compare it with five widely used baseline models: LSTM, Bi-LSTM, GRU, iTransformer, and S-Mamba. These models represent a diverse set of approaches for time-series forecasting, ranging from traditional recurrent architectures to advanced state space models.

\begin{itemize}
\item \textbf{LSTM (Long Short-Term Memory) \citep{seabe2023forecasting}:} Configured with 3 layers and a hidden size of 100. LSTM is a popular recurrent neural network architecture and is chosen for its ability at capturing long-term dependencies in sequential data. It is widely used in financial forecasting tasks.

\item \textbf{Bi-LSTM (Bidirectional LSTM) \citep{seabe2023forecasting}:} Configured with 3 layers and a hidden size of 100. The bidirectional setup allows the model to learn both forward and backward temporal dependencies, potentially capturing richer patterns and contextual dependencies in the data.

\item \textbf{GRU (Gated Recurrent Unit) \citep{seabe2023forecasting}:} Uses 3 layers and a hidden size of 100. The GRU model is a variant of LSTM that achieves comparable performance with fewer parameters. It is included as a lightweight alternative for evaluating the trade-off between accuracy and model complexity.

\item \textbf{iTransformer\citep{liu2023itransformer}:} 
Configured with \texttt{d\_model} 128, \texttt{n\_heads} 16, \texttt{d\_ff} 128, and \texttt{e\_layers} 2. iTransformer is a transformer-based model specifically designed for time-series forecasting and is included to assess the effectiveness of advanced attention-based architectures for Bitcoin price prediction.

\item \textbf{S-Mamba \citep{wang2024mamba}:} Configured with \texttt{d\_model} 128, \texttt{d\_state} 32, \texttt{d\_ff} 128, 0.1 \texttt{dropout}, and \texttt{e\_layers} 2. S-Mamba has not been previously applied to Bitcoin price prediction or other cryptocurrency datasets. We include S-Mamba to assess \method{}’s performance against a state-of-the-art model with similar underlying principles.

\item \textbf{\method{}:} Configured with 3 C-Blocks, each containing 4 CMBlocks. The sequence lengths for the C-Blocks are 14, 16, and 32, respectively, with a state dimension (\texttt{d\_state}) of 64 for the Mamba blocks.

\end{itemize}

In the experiments, each model receives data from the previous 14 days as input to predict the \textit{close} value for the following day. To ensure consistency across models, we use the Adam optimizer \citep{arXiv17_Adam} with RMSE loss as the loss function. All models are trained with a batch size of 32. Additionally, we employ a learning rate scheduler and weight decay to mitigate overfitting and tune these hyperparameters for each model. Early stopping is applied, and the model checkpoint with the best validation loss is selected to avoid overfitting.

All models use the same train-validation-test split to enable fair comparisons. To avoid data contamination, our predictions in each data split period start from 14 days after the start of that period. This means that, for example, for validation samples, our prediction starts with a 14-day delay to avoid using training samples in the input. This gap ensures that each set remains completely isolated, preventing any overlap and enabling a more accurate assessment of each model’s performance on unseen data. Additionally, we conduct experiments under two different setups: with and without trading volume as a feature. This allows us to analyze the impact of volume on prediction accuracy and assess the robustness of the models under varying input configurations.

For all baselines, we use hyperparameters that align with their respective best practices, ensuring a fair comparison. Additionally, to maintain reproducibility, we set a fixed random seed for all experiments.

\subsection{Evaluation}

\begin{figure*}[t]
	\centering
	\subfloat[\method{}]{
            \includegraphics[width=0.32\textwidth]{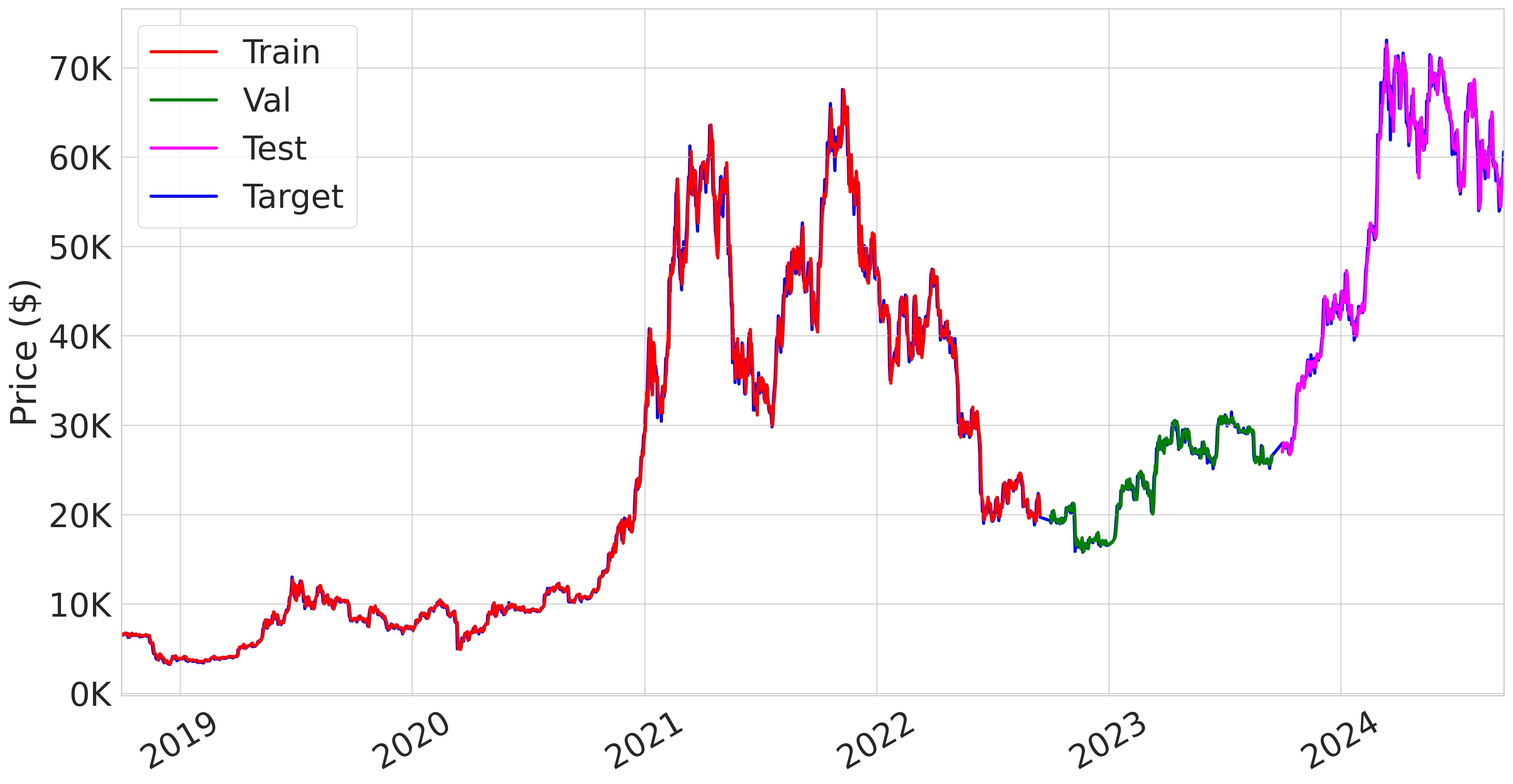}
        }
        \subfloat[LSTM]{
            \includegraphics[width=0.32\textwidth]{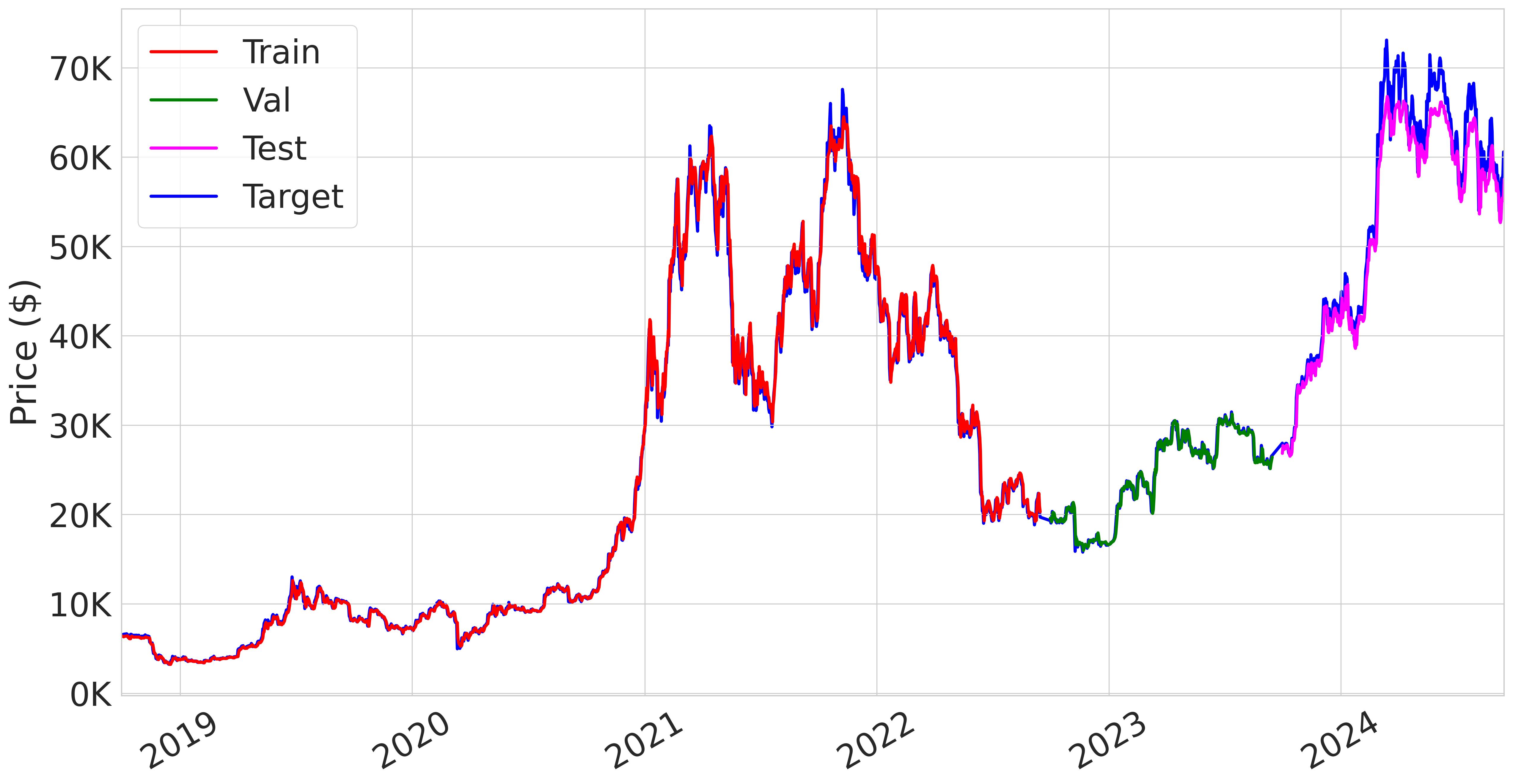}
        }
        \subfloat[Bi-LSTM]{
            \includegraphics[width=0.32\textwidth]{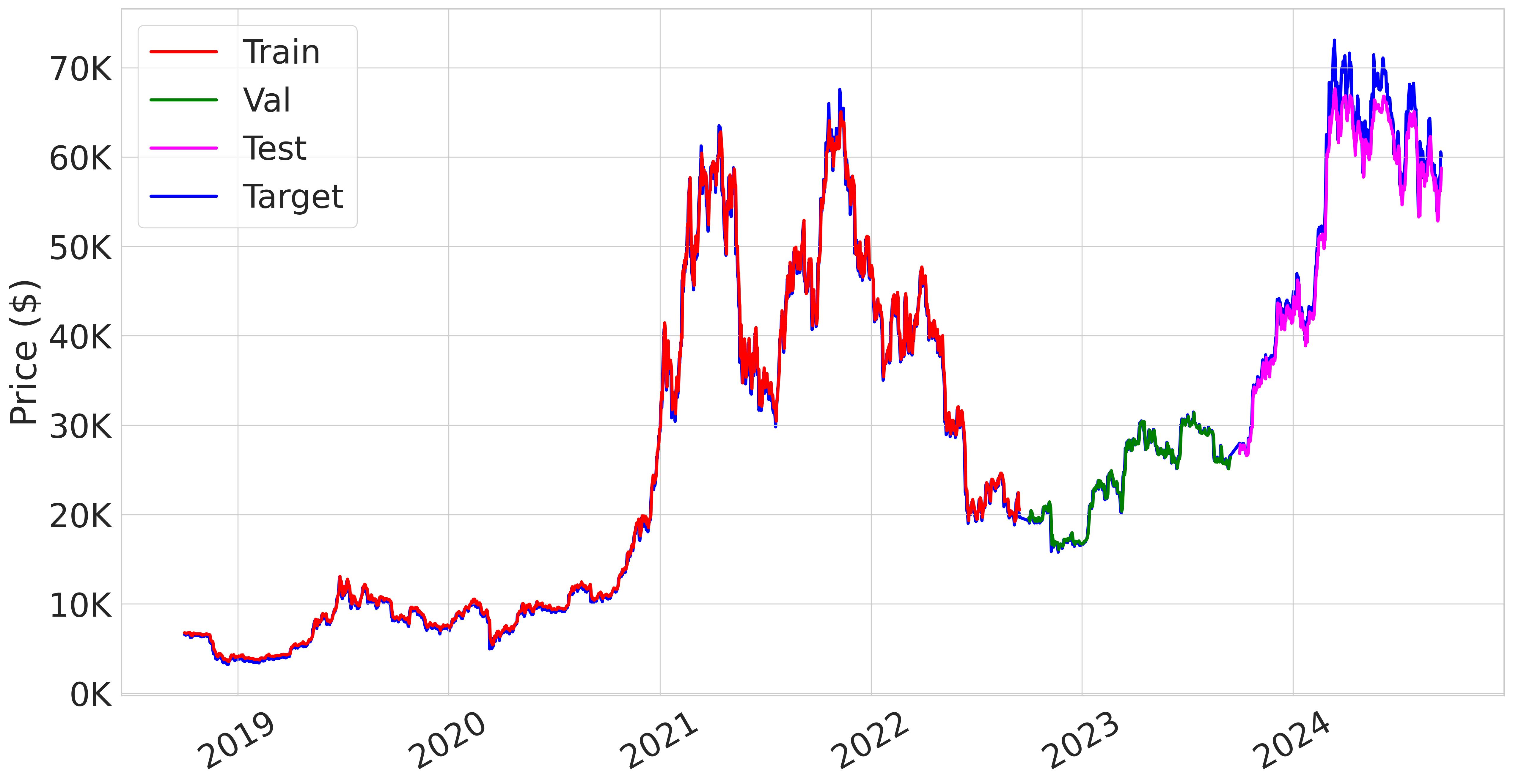}
        }\\
        \subfloat[GRU]{
            \includegraphics[width=0.32\textwidth]{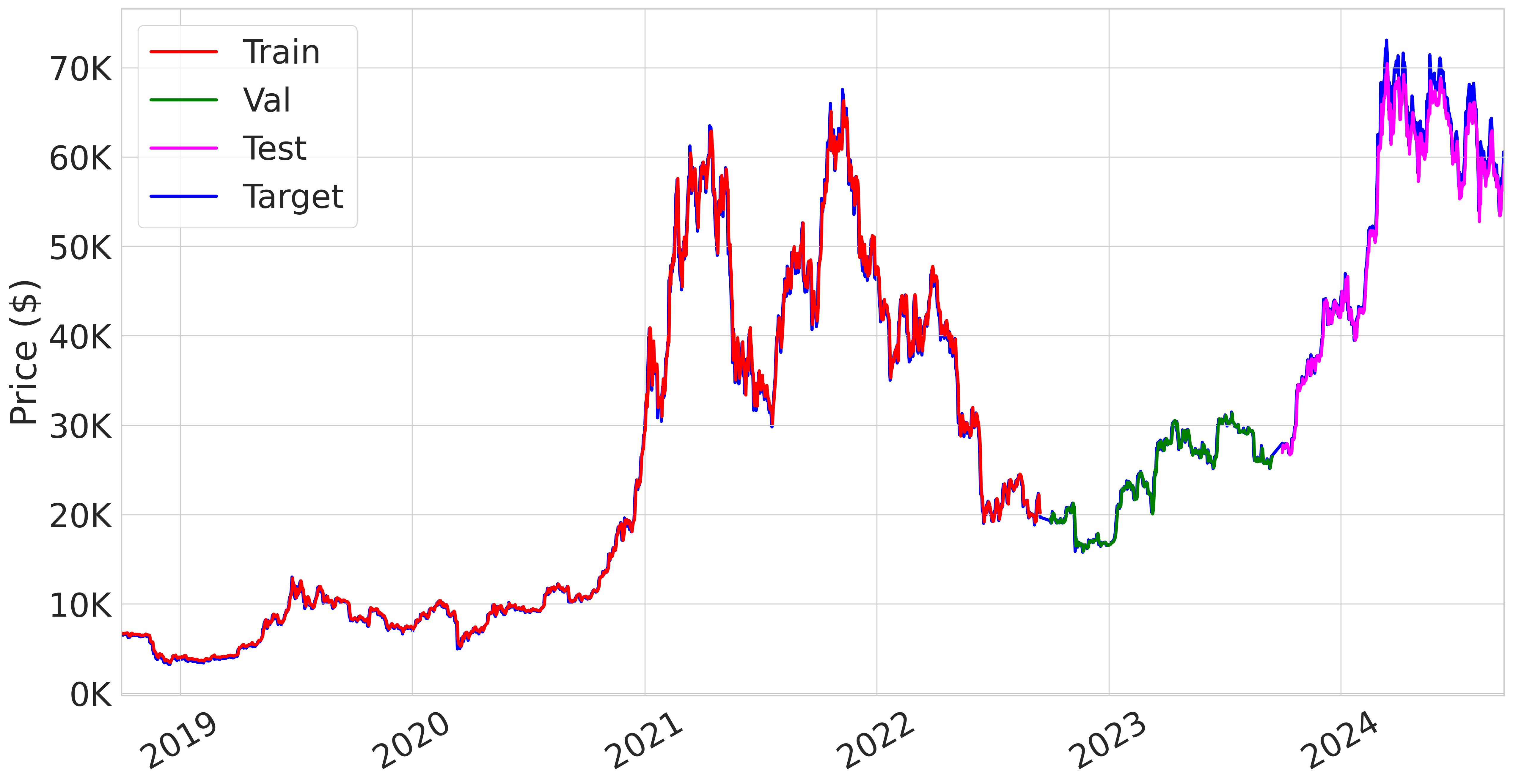}
        }
        \subfloat[iTransformer]{
            \includegraphics[width=0.32\textwidth]{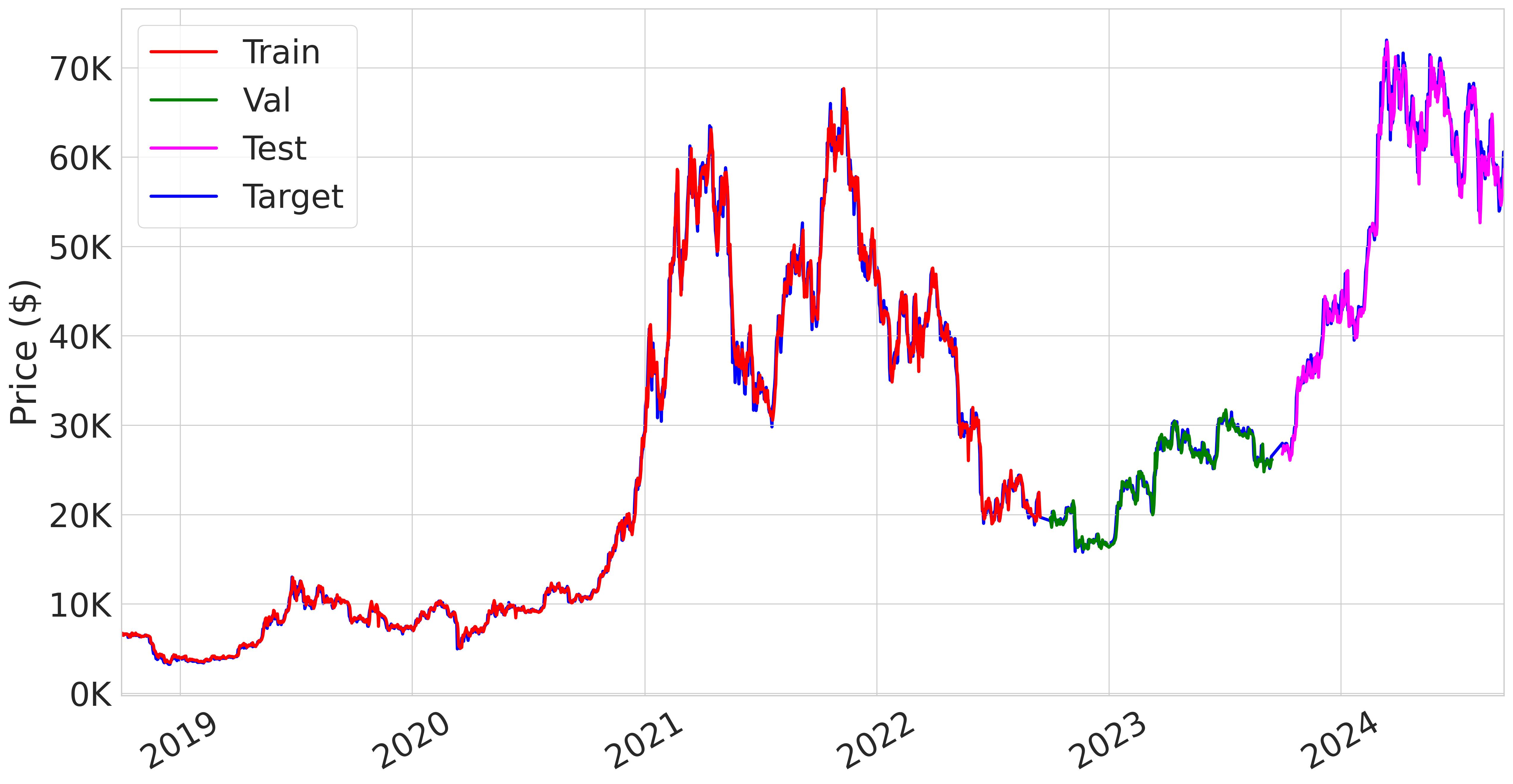}
        }
        \subfloat[S-Mamba]{
            \includegraphics[width=0.32\textwidth]{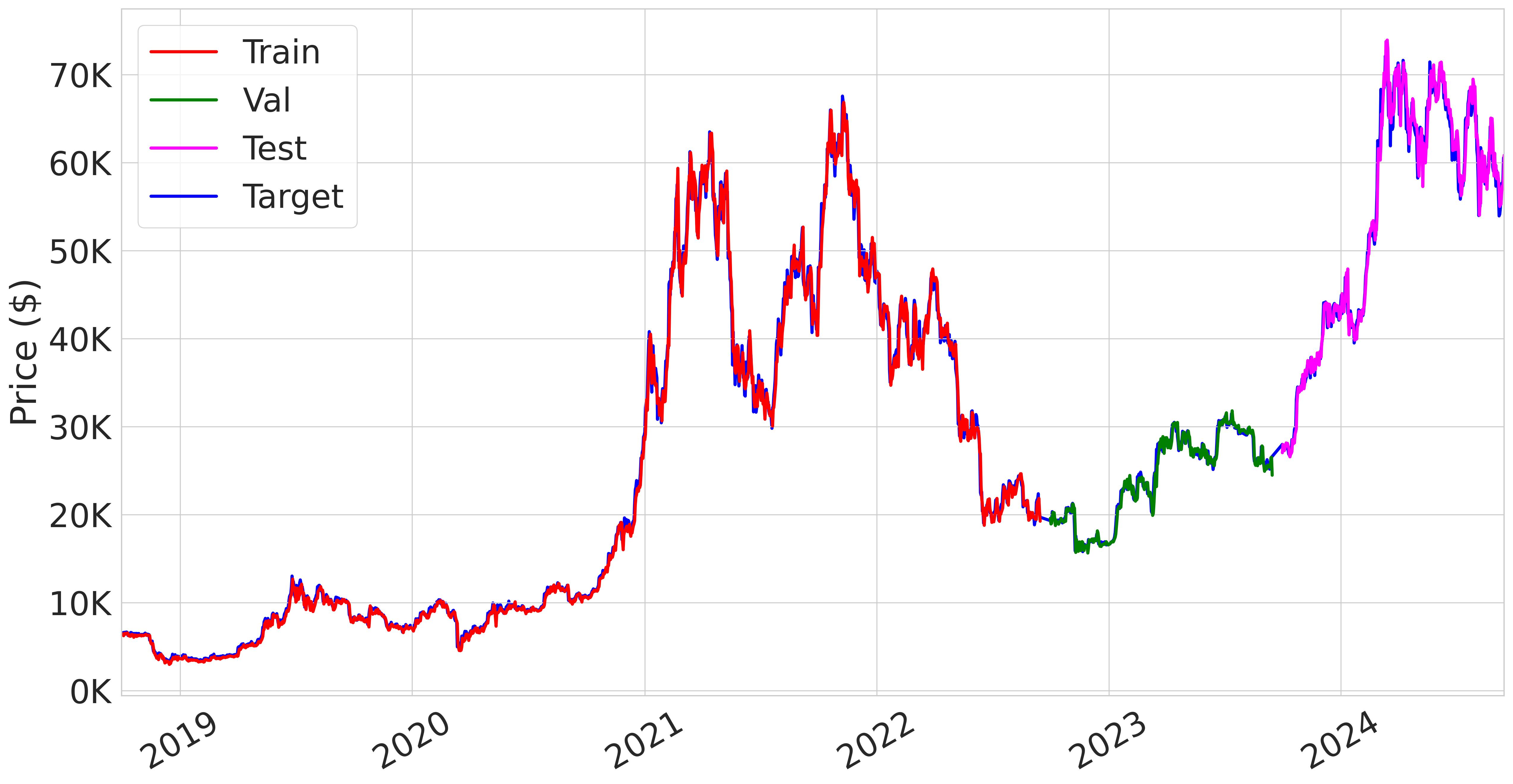}
        }
        \caption{Forecasting results for all models (a) \method{}, (b) LSTM, (c) Bi-LSTM, (d) GRU, (e) iTransformer, and (f) S-Mamba on the training, validation, and test sets without using volume data. Non-Mamba models struggle to capture large price fluctuations, often underperforming during periods of high volatility.}
	\label{fig:no_vol}
\end{figure*}

The results of our experiments, conducted with and without trading volume on the test data, are summarized in Table \ref{table:results}. These results show that \method{} consistently outperforms all baseline models, including LSTM, Bi-LSTM, GRU, iTransformer, and S-Mamba, across all evaluation metrics. This demonstrates \method{}'s superior ability to capture the complex dynamics of Bitcoin price movements with high generalization. Notably, \method{}-v, the volume-inclusive variant, achieves the best performance with an RMSE of 1598.1, a MAPE of 2.034, and an MAE of 1120.7. Even without volume, \method{} surpasses the volume-inclusive versions of all baseline models, highlighting its robustness.

Among the baselines, S-Mamba performs competitively, particularly in the volume-inclusive setup, with an RMSE of 1651.6. This highlights the strength of advanced state space models in capturing long-range dependencies. However, \method{}'s tailored architecture further improves upon these results, demonstrating the benefits of its design for financial time-series forecasting. The iTransformer baseline performs better than traditional recurrent methods in most metrics. However, it still falls short of the performance achieved by Mamba-based models, suggesting that while attention mechanisms are effective, SSMs offer a more suitable framework for handling long-range dependencies in highly volatile financial data. Furthermore, despite having relatively low accuracy in the volume-exclusive setup, Bi-LSTM and LSTM show significant improvements when volume data is included, whereas GRU sees minimal gains, indicating that the impact of volume depends on the model architecture. 

Overall, the results confirm \method{}'s effectiveness in both setups, with the inclusion of volume data enhancing prediction accuracy across most models. This highlights the importance of trading volume as a feature in capturing market dynamics, particularly in highly volatile settings like Bitcoin price forecasting.

\begin{figure*}[ht]
	\centering
	\subfloat[\method{}]{
            \includegraphics[width=0.32\textwidth]{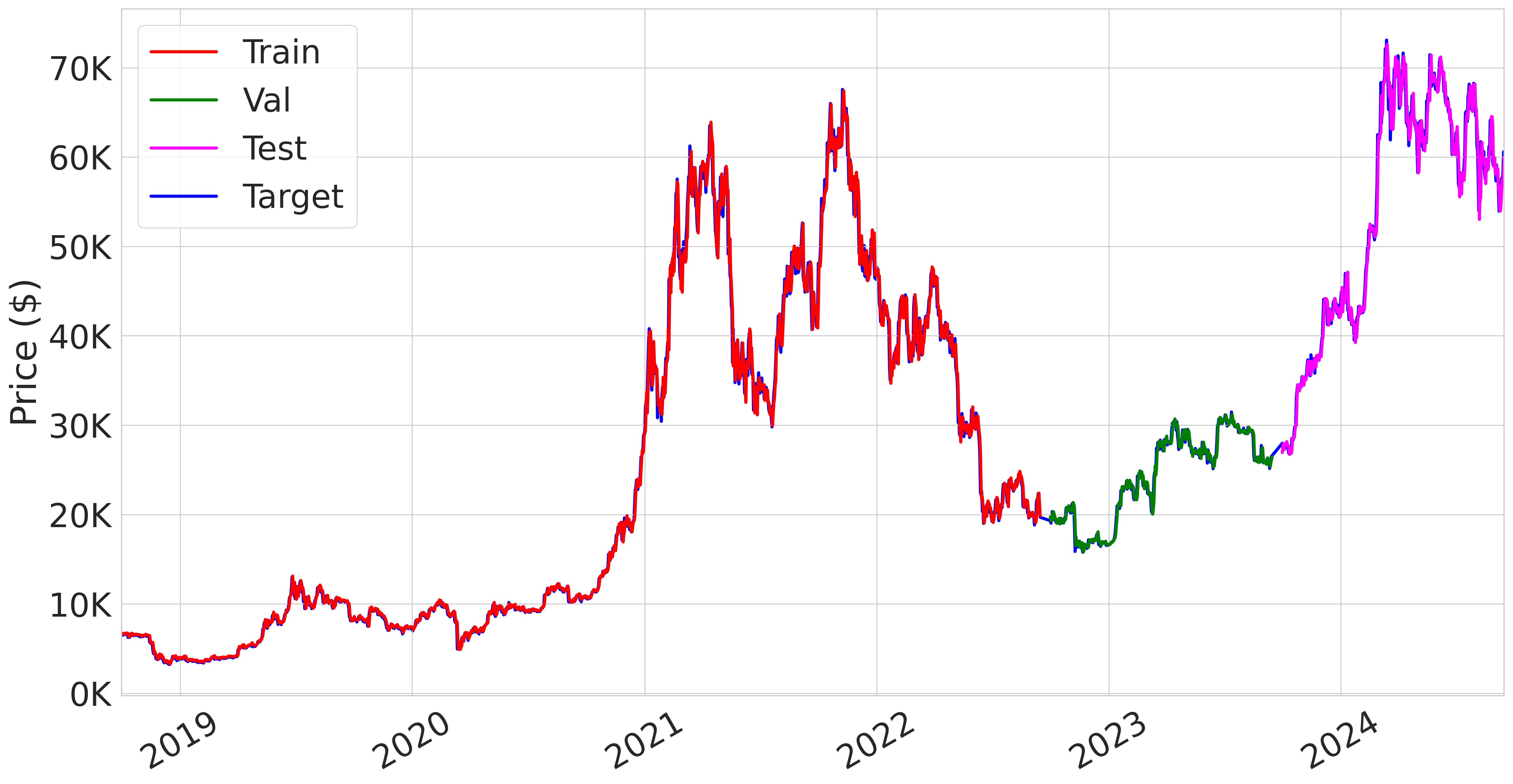}
        }
        \subfloat[LSTM]{
            \includegraphics[width=0.32\textwidth]{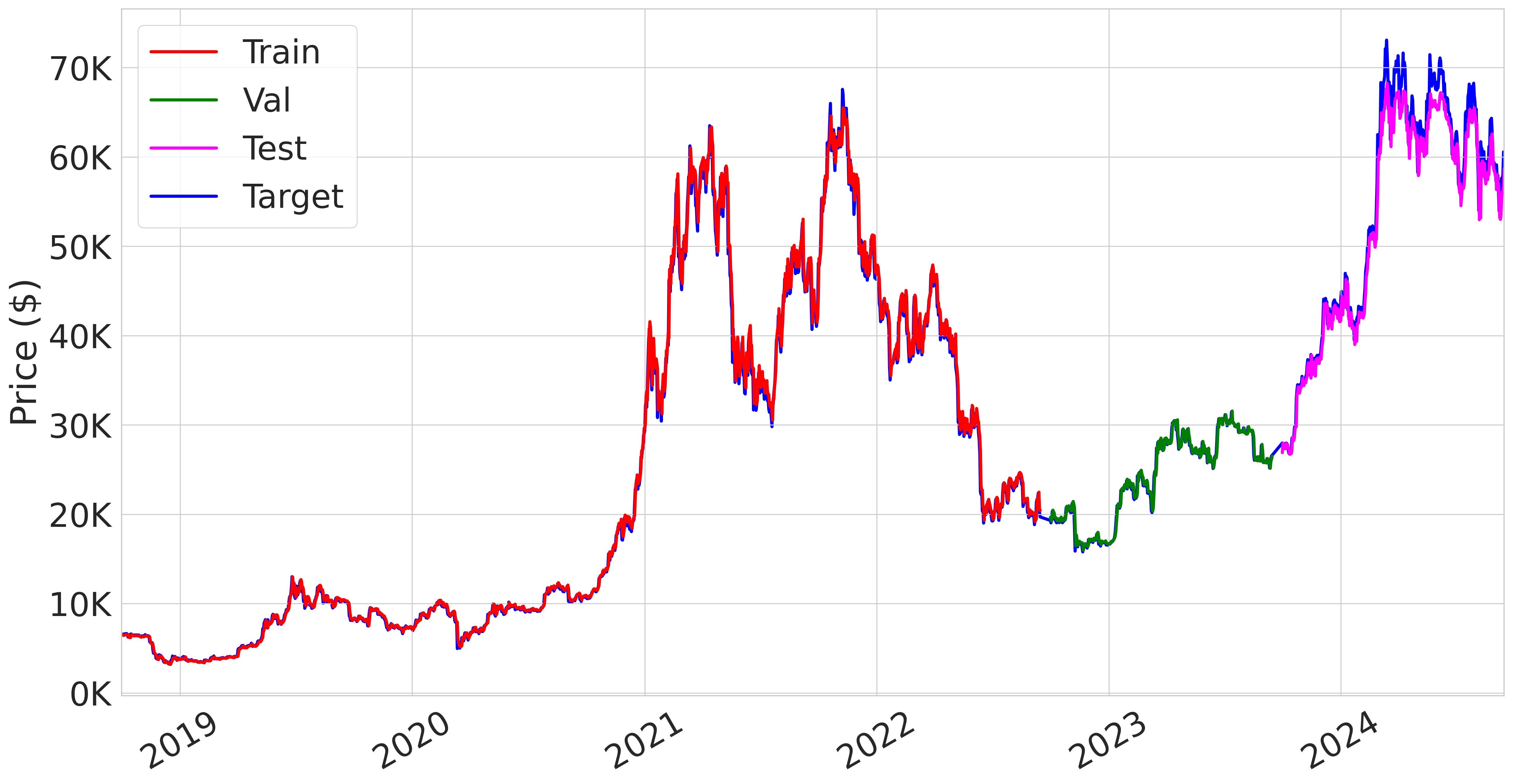}
        }
        \subfloat[Bi-LSTM]{
            \includegraphics[width=0.32\textwidth]{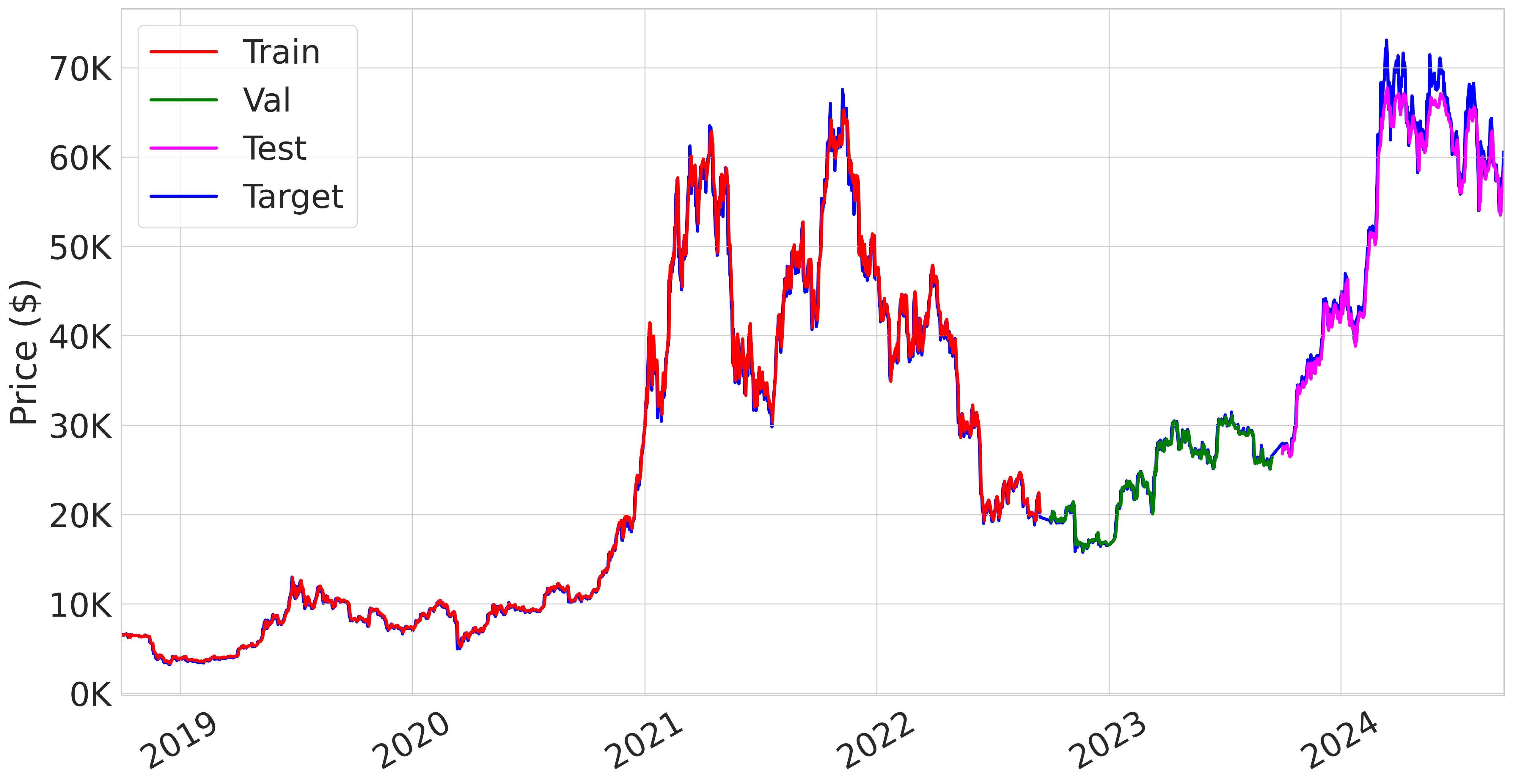}
        }\\
        \subfloat[GRU]{
            \includegraphics[width=0.32\textwidth]{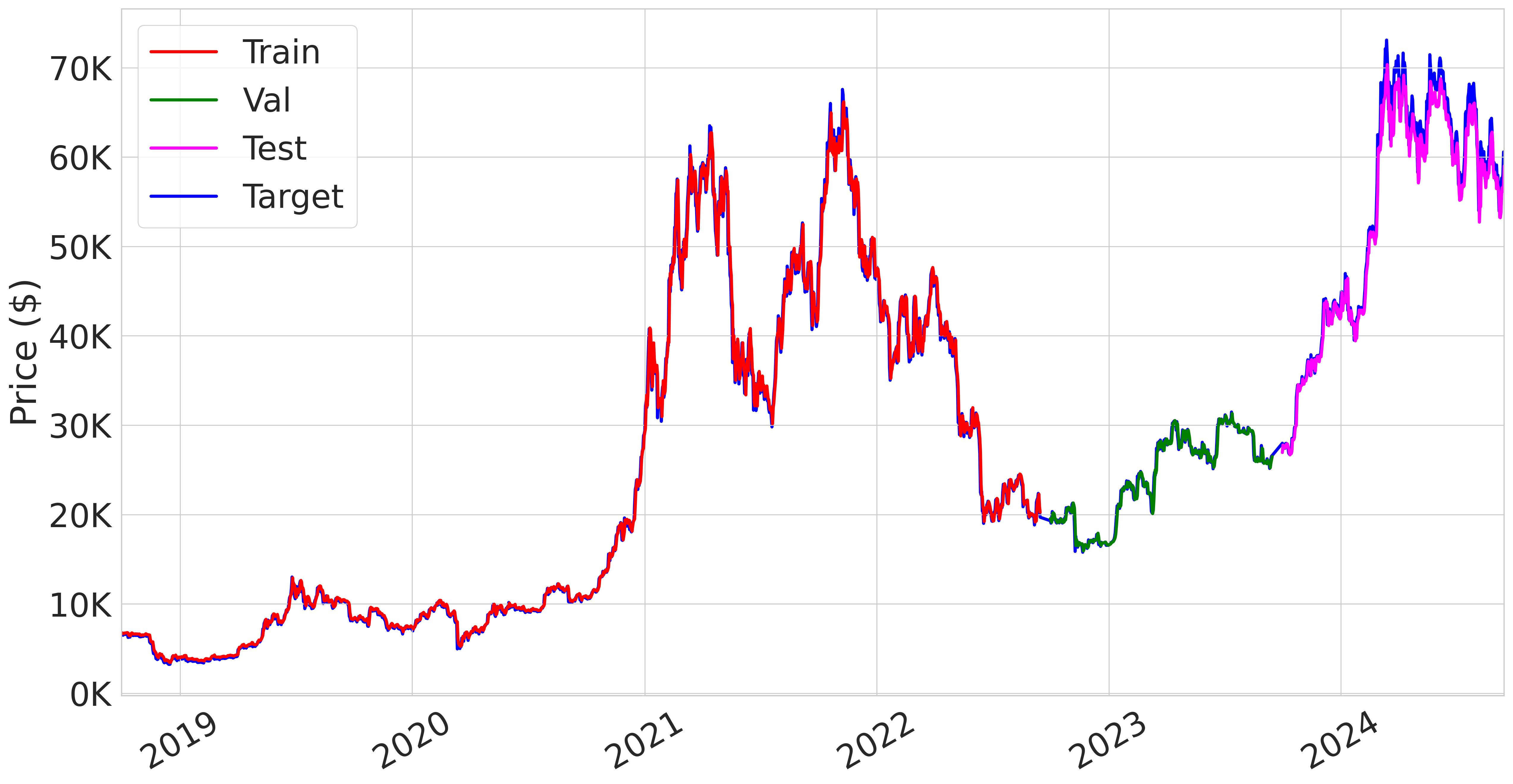}
        }
        \subfloat[iTransformer]{
            \includegraphics[width=0.32\textwidth]{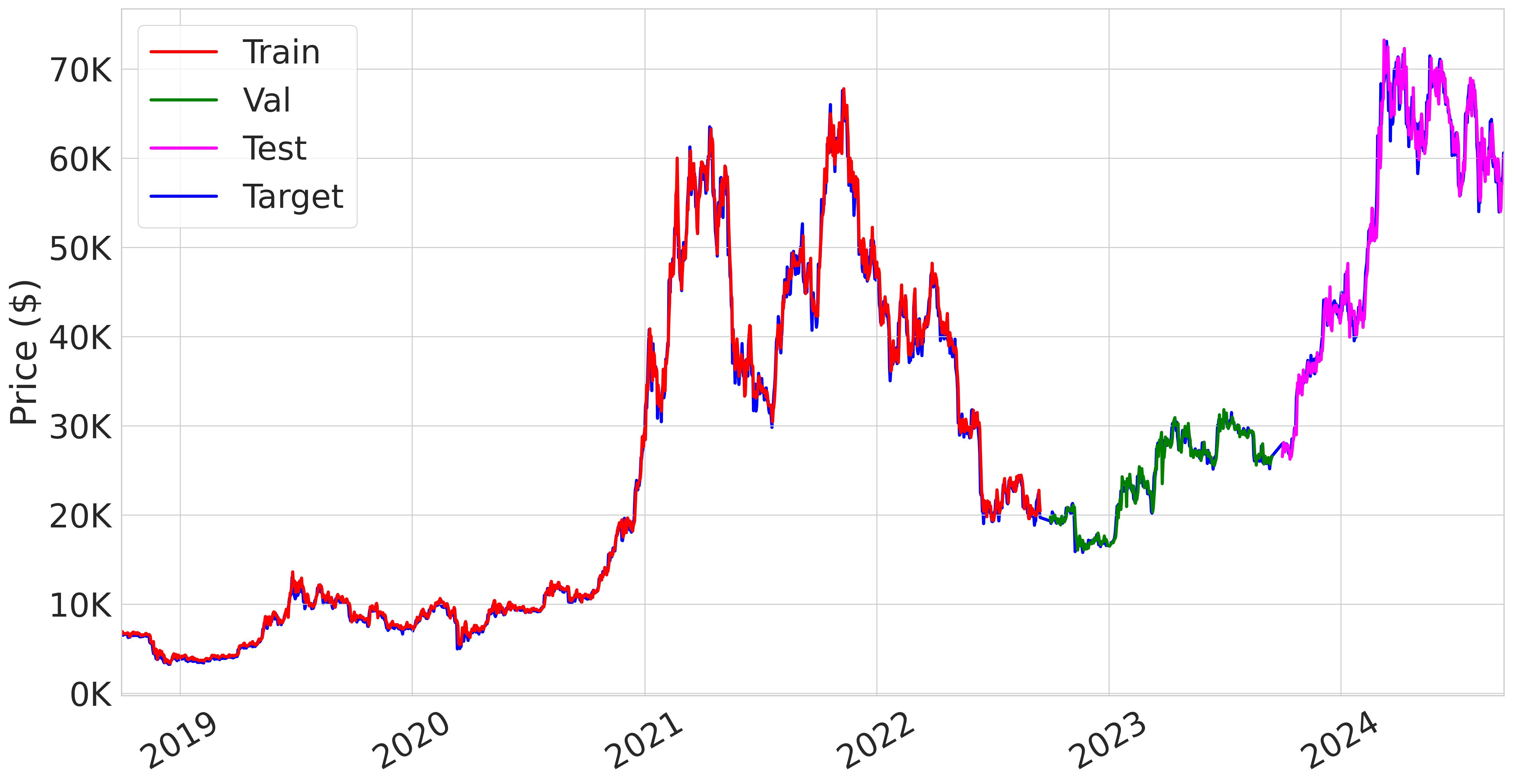}
        }
        \subfloat[S-Mamba]{
            \includegraphics[width=0.32\textwidth]{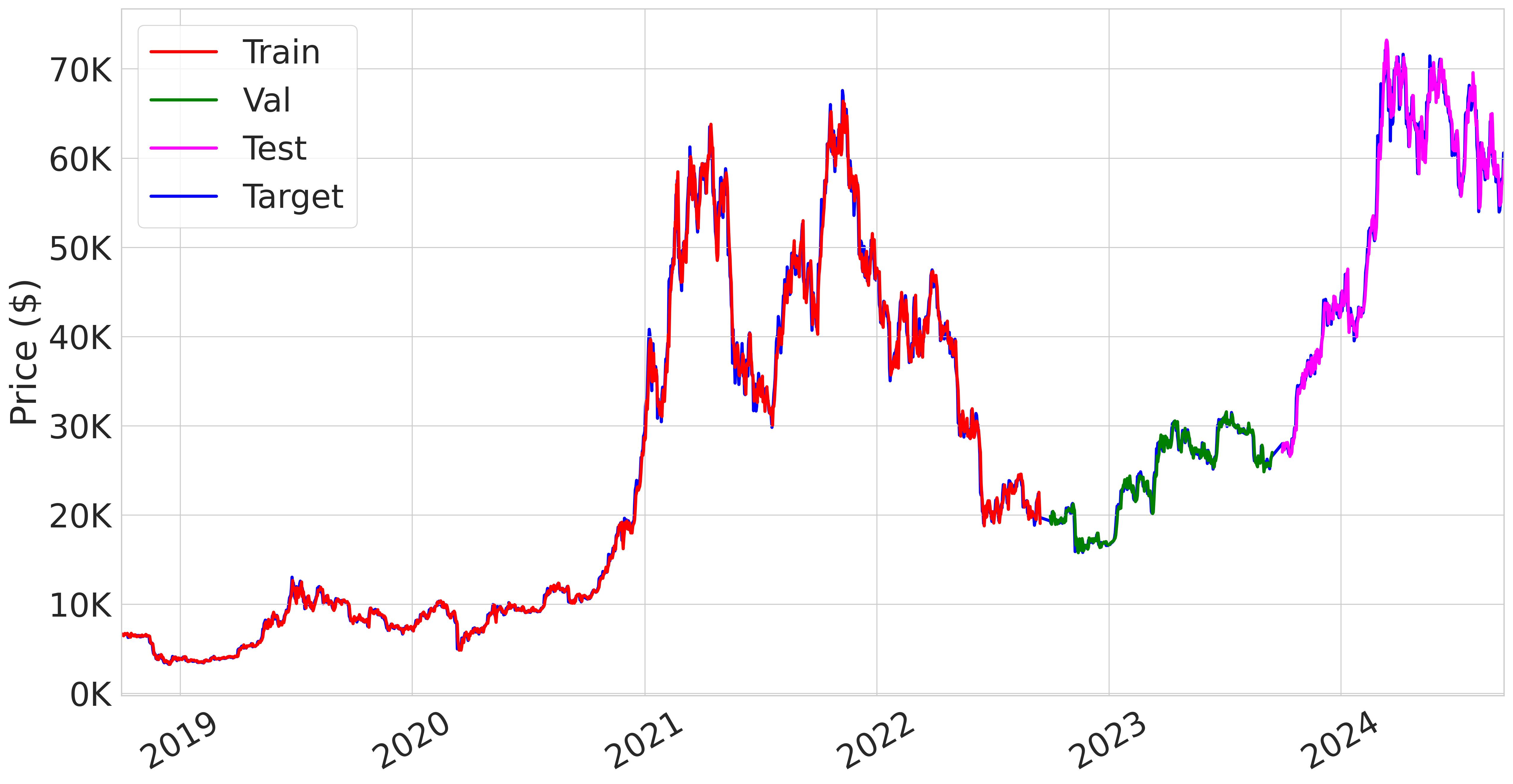}
        }
        \caption{Forecasting results for all models (a) \method{}, (b) LSTM, (c) Bi-LSTM, (d) GRU, (e) iTransformer, and (f) S-Mamba on the training, validation, and test sets with volume data included as an additional feature. While volume data helps non-Mamba models slightly, they still face challenges in accurately predicting large price changes, especially during volatile periods.}
	\label{fig:vol}
\end{figure*}

Figures \ref{fig:no_vol} and \ref{fig:vol} illustrate the forecasting results of all models on the training, validation, and test sets, without and with volume data, respectively. While all models perform well on the training set, the non-Mamba baselines, struggle to maintain accuracy on the test set during volatile periods, showing significant divergence from actual values. S-Mamba performs better than other baselines but still falls short in capturing large price fluctuations. In contrast, \method{} consistently tracks actual price trends, demonstrating superior generalization and robustness across both setups. The inclusion of volume data further improves the performance of most models, with \method{} leveraging this additional information to achieve the most accurate predictions, particularly during periods of high volatility.

\subsection{Efficiency}
Table~\ref{table:results} reports the number of parameters for each model alongside their regression metrics, allowing a direct comparison of predictive performance and model complexity. With only 136k parameters, \method{} has the smallest footprint among all baselines, significantly outperforming much larger models like Bi-LSTM (569k) and S-Mamba (330k). Notably, even LSTM and GRU, with 204k and 153k parameters, respectively, exhibit higher complexity than \method{}. This compact design enables \method{} to capture essential temporal patterns without relying on a large parameter count.

\begin{table}[!ht]
    \centering
    \caption{Results on test data, \textit{-v} shows including volume in the data}
    \label{table:results}
    \renewcommand{\arraystretch}{1.3} 
    \resizebox{0.55\textwidth}{!}{
    \begin{tabular}{lcccccc}
        \toprule
        \textbf{Method} & \textbf{RMSE} & \textbf{MAPE} & \textbf{MAE} && \textbf{Parameters}\\
        \midrule
        LSTM \citep{seabe2023forecasting} & 2672.7 & 3.609 & 2094.3 && 204k\\
        Bi-LSTM \citep{seabe2023forecasting} & 2325.6 & 3.072 & 1778.8 && 569k\\
        GRU \citep{seabe2023forecasting} & 1892.4 & 2.385 & 1371.2 && 153k\\
        iTransformer \citep{liu2023itransformer} & 1826.9 & 2.460 & 1334.3  && 201k\\
        S-Mamba \citep{wang2024mamba} & 1717.4 & 2.248 & 1239.9 && 330k\\
        \method & \textbf{1713.0} & \textbf{2.171} & \textbf{1200.9} && \textbf{136k}\\
        \midrule
        LSTM-v & 2202.1 & 2.896 & 1668.9 && 204k\\
        Bi-LSTM-v & 2080.2 & 2.738 & 1562.5 && 569k\\
        GRU-v & 1978.0 &  2.526 & 1454.3 && 153k\\
        iTransformer-v & 1905.9 & 2.540 & 1387.47 && 201k\\
        S-Mamba-v & {1651.6} & 2.215 & 1209.7 && 330k \\
        \method-v & \textbf{1598.1} & \textbf{2.034}  & \textbf{1120.7} && \textbf{136k}\\
        \bottomrule
    \end{tabular}
    }
\end{table} 

Despite its smaller size, \method{} achieves state-of-the-art results, demonstrating an effective trade-off between efficiency and accuracy. The reduced model complexity lowers memory and computational demands, minimizing overfitting and making \method{} more suitable for deployment in resource-constrained environments. For instance, compared to S-Mamba, which requires over twice the number of parameters, \method{} delivers superior predictive accuracy with substantially less complexity.

In addition to its compact architecture and strong performance, \method{} is highly efficient in both training and inference, making it well-suited for real-time applications. On a single NVIDIA Quadro RTX 5000 GPU, training CryptoMamba takes only 29 minutes and consumes 516 MB of GPU memory. For inference, \method{} achieves a speed of 1.17\,ms per sample with a batch size of 128 on the same GPU. These results further highlight \method{}’s suitability for deployment in latency-sensitive and resource-constrained environments.

%% file: sections/6_application.tex
\section{Application in Real-World Trading}\label{sec:application}

In this section, we evaluate the practical utility of \method{} by applying its predictions in a real-world trading scenario. Using three trading algorithms, we simulate trading with an initial balance of 100 dollars at the beginning of the validation and test period. The goal is to compute the final net worth and evaluate the risk-adjusted performance of different models. To assess risk, we use the Maximum Drawdown (MDD), a widely-used metric that measures the maximum observed loss from a peak to a trough over the trading horizon (see Appendix~\ref{sec:metrics} for details).

\begin{minipage}[c]{0.48\textwidth}
\begin{algorithm}[H]
\caption{Vanilla Trading Algorithm}
\label{alg:vanilla_trading}
\begin{algorithmic}[1]
\REQUIRE Prediction, Today's Price, threshold
\STATE $x \gets \text{Today's price}$
\STATE $y \gets \text{Prediction}$
\STATE $d \gets \left| \frac{x - y}{x} \right|$
\IF{$d \geq \text{threshold}$}
    \IF{$x > y$}
        \STATE Sell all of your shares
    \ELSE
        \STATE Buy with all of your money
    \ENDIF
\ELSE
    \STATE Do not buy or sell
\ENDIF
\end{algorithmic}
\end{algorithm}
\end{minipage}\hfill
\begin{minipage}[c]{0.48\textwidth}
\begin{algorithm}[H]
\label{alg:two}
\caption{Smart Trading Algorithm}
\label{alg:smart_trading}
\begin{algorithmic}[1]
\REQUIRE Prediction, Today's Price, risk
\STATE $x \gets \text{Today's price}$
\STATE $y \gets \text{Prediction}$
\STATE $y_{\max} \gets \left(1 + \frac{\text{risk}}{100}\right)y$
\STATE $y_{\min} \gets \left(1 - \frac{\text{risk}}{100}\right)y$
\IF{$x \geq y$}
    \IF{$x \geq y_{\max}$}
        \STATE Sell all of your shares
    \ELSE
        \STATE Sell $\frac{x - y}{y_{\max} - y}$ of your shares
    \ENDIF
\ELSE
    \IF{$x \leq y_{\min}$}
        \STATE Buy with all of your money
    \ELSE
        \STATE {Buy with $\frac{y - x}{y - y_{\min}}$ of your money}
    \ENDIF
\ENDIF
\end{algorithmic}
\end{algorithm}\end{minipage}

To evaluate model predictions, we use three trading algorithms: Vanilla, Smart, and Extended Smart. Algorithm \ref{alg:vanilla_trading} describes the \textbf{Vanilla Trading Algorithm}, which makes simple buy or sell decisions based on the predicted and actual prices. At its core, the algorithm computes the ratio of the absolute difference between the predicted price and today's price to today's price. This ratio, referred to as the change ratio $d$, determines whether any trading action is required. If the change ratio is below a predefined threshold (0.01 in our case), the algorithm refrains from making any trades. This safeguard is introduced to account for transaction fees, which can render small trades unprofitable, especially when the predicted price change is minor. However, if the change ratio meets or exceeds the threshold, the algorithm takes decisive action. 

Algorithm \ref{alg:smart_trading}, the \textbf{Smart Trading Algorithm}, introduces a risk-aware approach by considering an interval around the predicted price, determined by a risk percentage (set to 2\% based on the validation MAPE of the models). This interval, defined by an upper bound ($y_{\text{max}}$) and a lower bound ($y_{\text{min}}$), accounts for prediction uncertainty and provides a range within which tomorrow's price is likely to fall. The algorithm makes decisions based on the position of today's price ($x$) relative to this interval. By dynamically adjusting trading actions based on this interval, the Smart Trading Algorithm balances risk and returns, offering a more adaptive and realistic trading strategy compared to the threshold-based Vanilla algorithm. 

The \textbf{Extended Smart Trading Algorithm} builds upon the Smart strategy by introducing support for short positions, allowing the agent to sell Bitcoin it does not own when the predicted price is expected to fall significantly. The strategy still operates based on the predicted price interval derived from the model’s risk percentage but includes a tunable cap on the maximum allowed short position (see Appendix~\ref{sec:extended_smart} for algorithm details). This addition enables more flexible trading decisions, particularly under highly volatile conditions, and offers a more comprehensive evaluation of model performance.

\begin{figure*}[t]
	\centering
        \subfloat[Vanilla - no volume]{
            \includegraphics[width=0.32\textwidth]{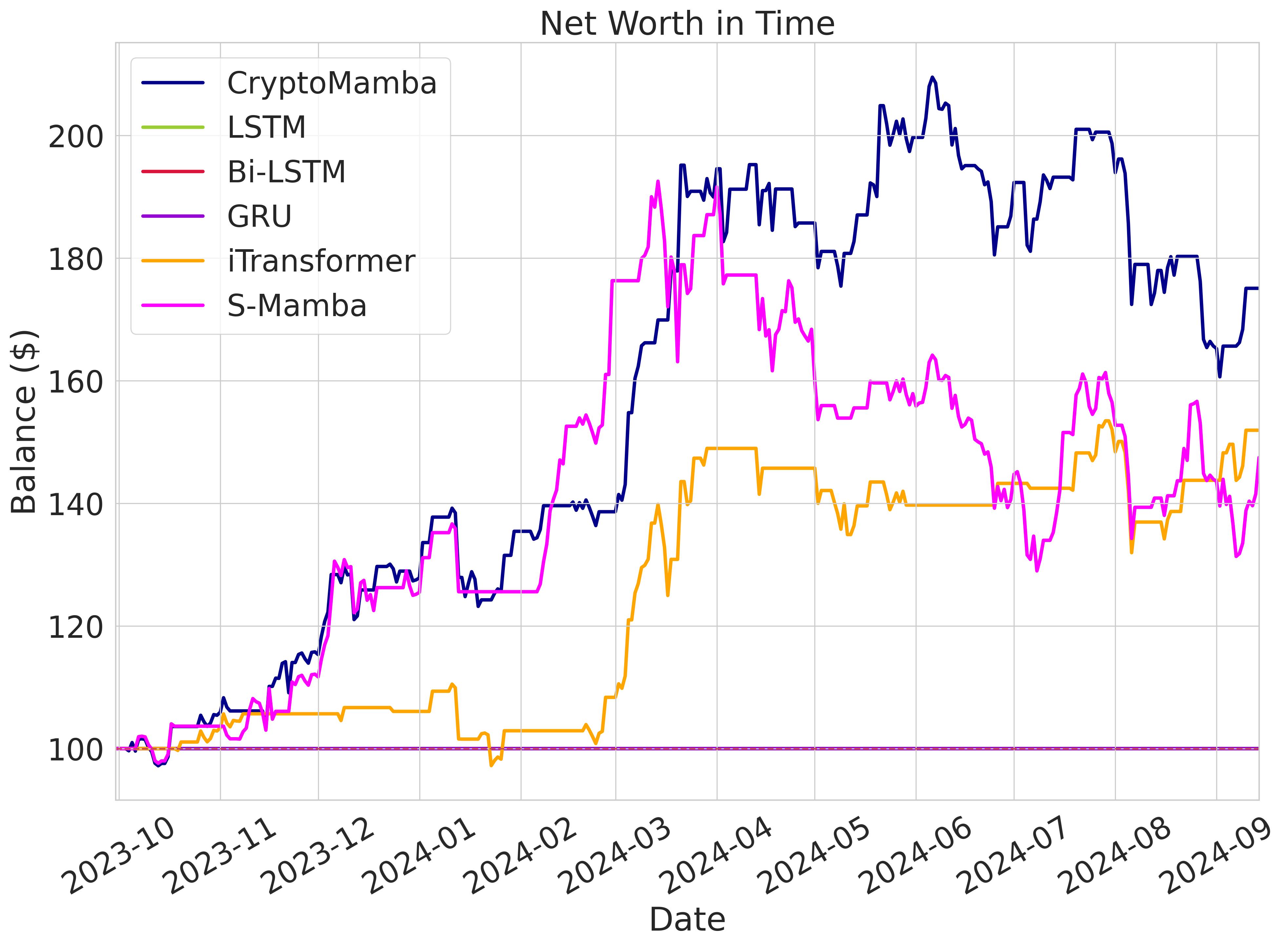}
        }
        \subfloat[Smart - no volume]{
            \includegraphics[width=0.32\textwidth]{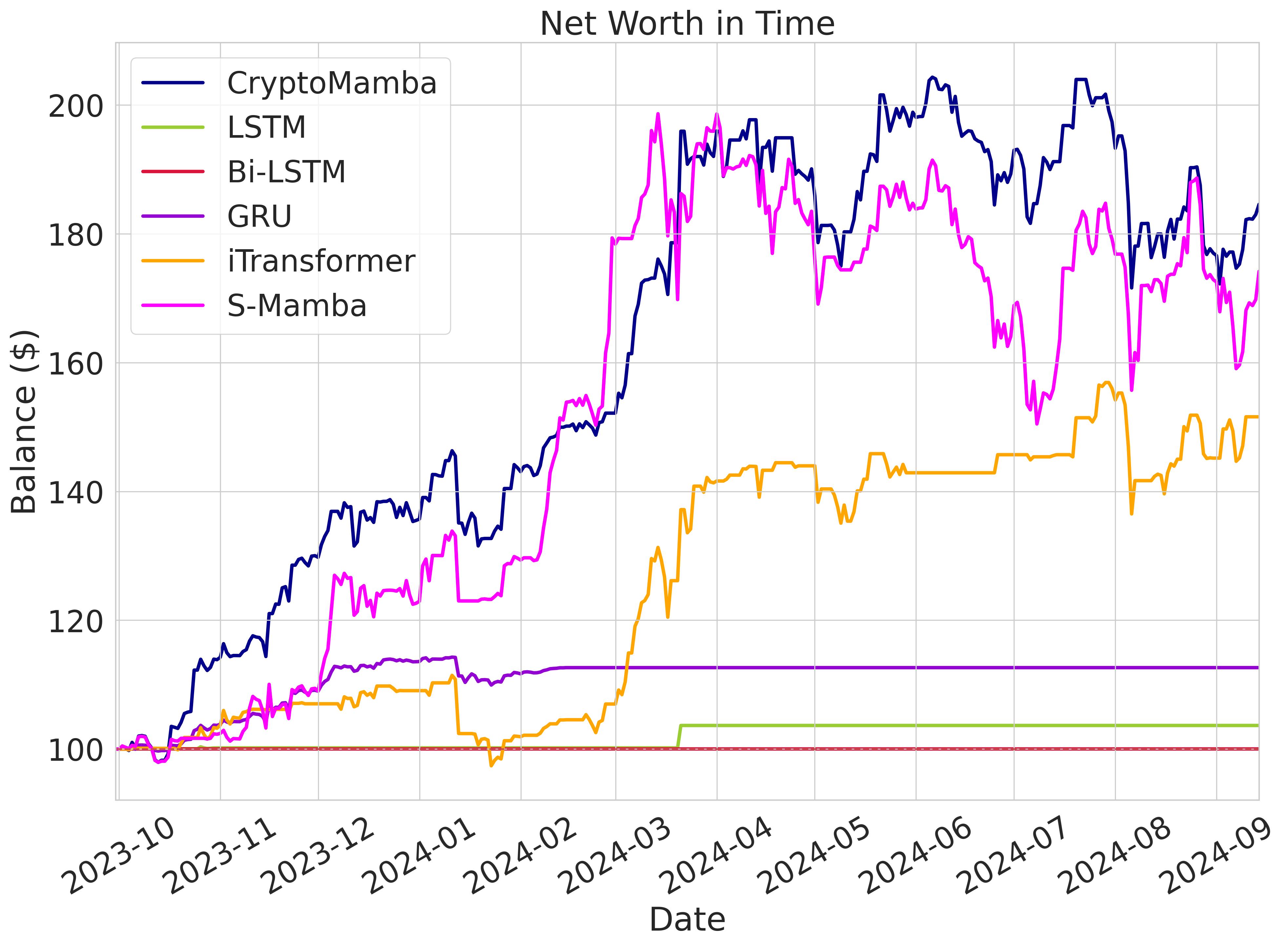}
        }
        \subfloat[Extended Smart - no volume]{
            \includegraphics[width=0.32\textwidth]{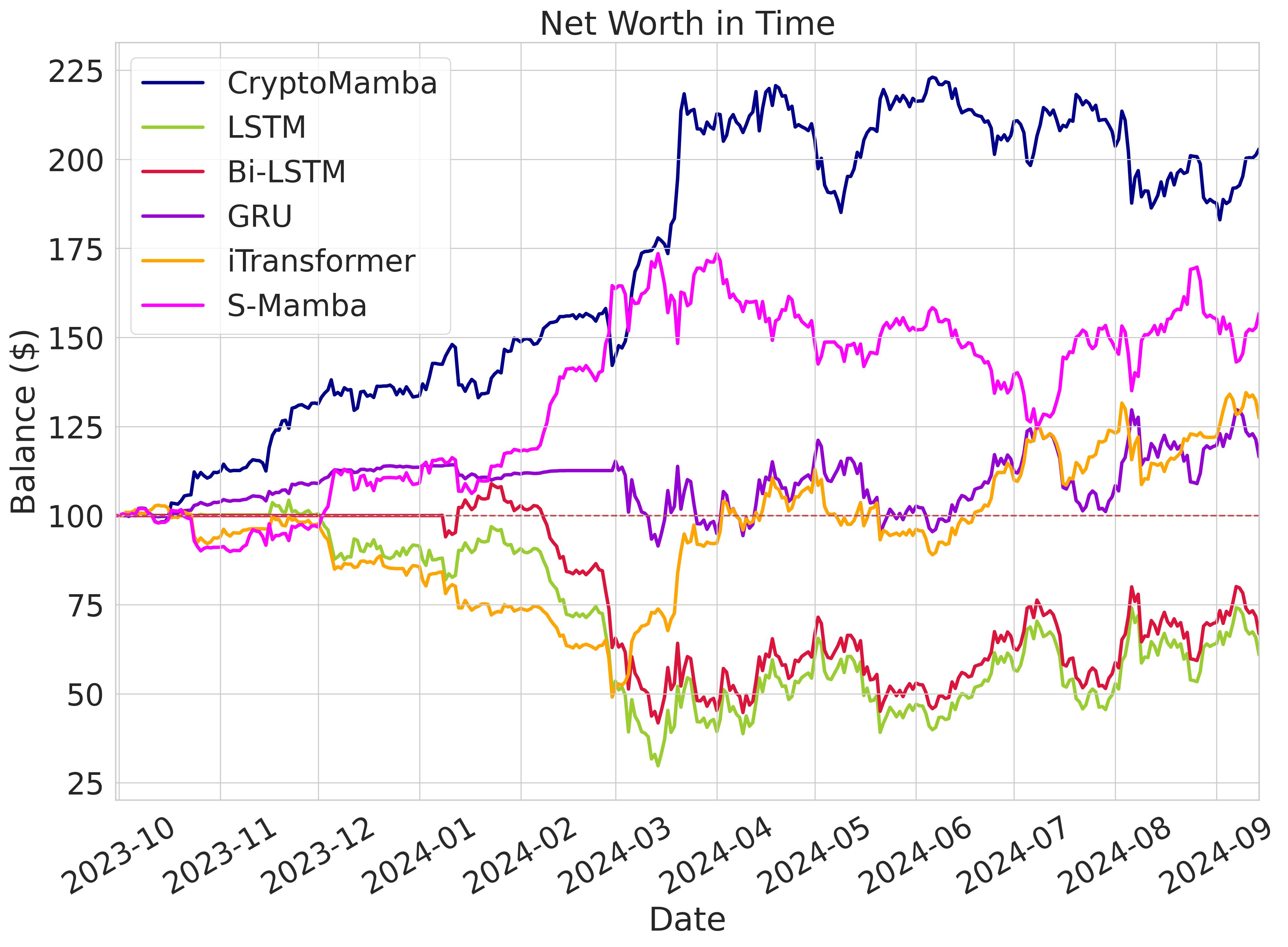}
        }
        
        \subfloat[Vanilla - with volume]{
            \includegraphics[width=0.32\textwidth]{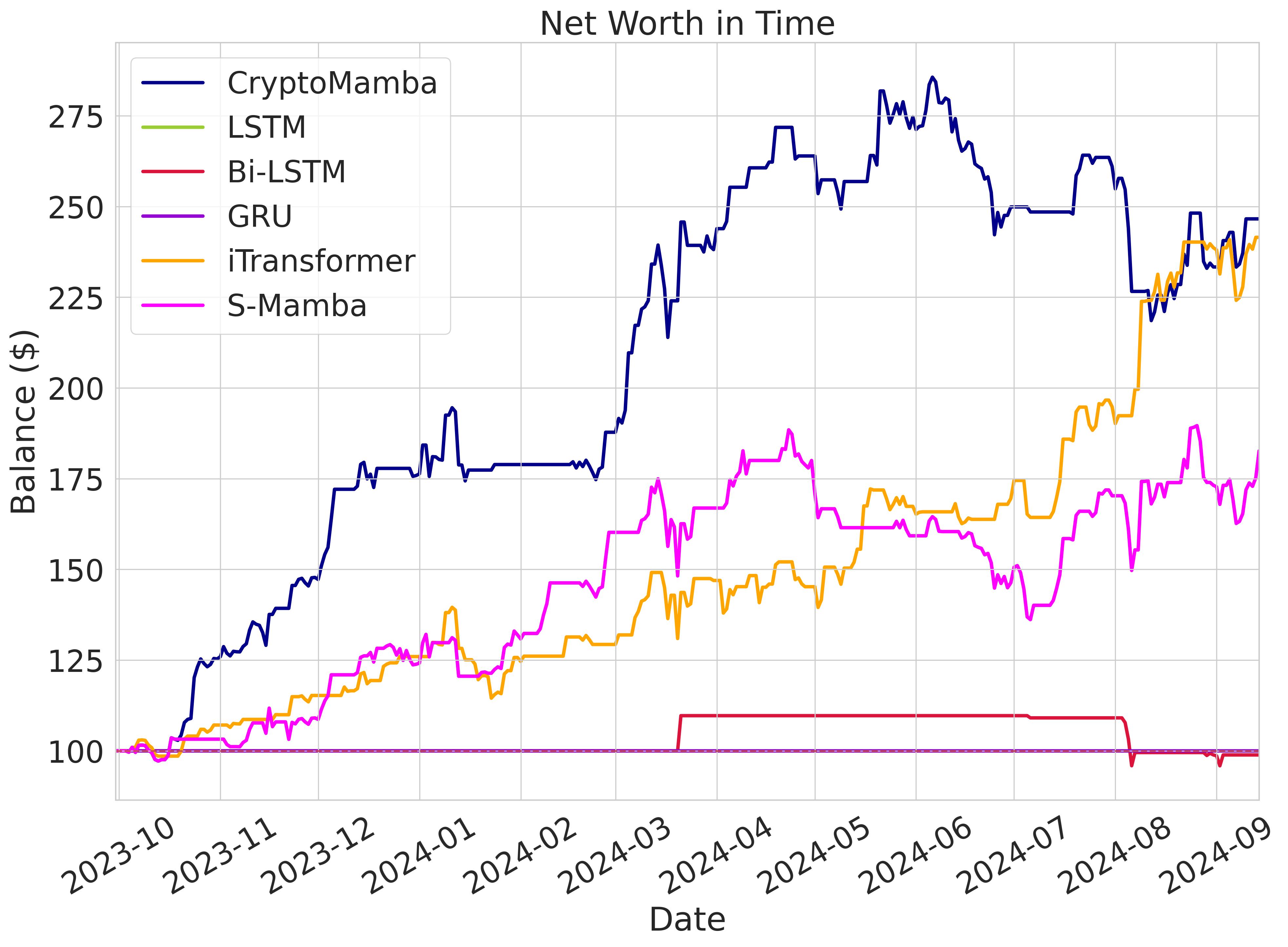}
        }
        \subfloat[Smart - with volume]{
            \includegraphics[width=0.32\textwidth]{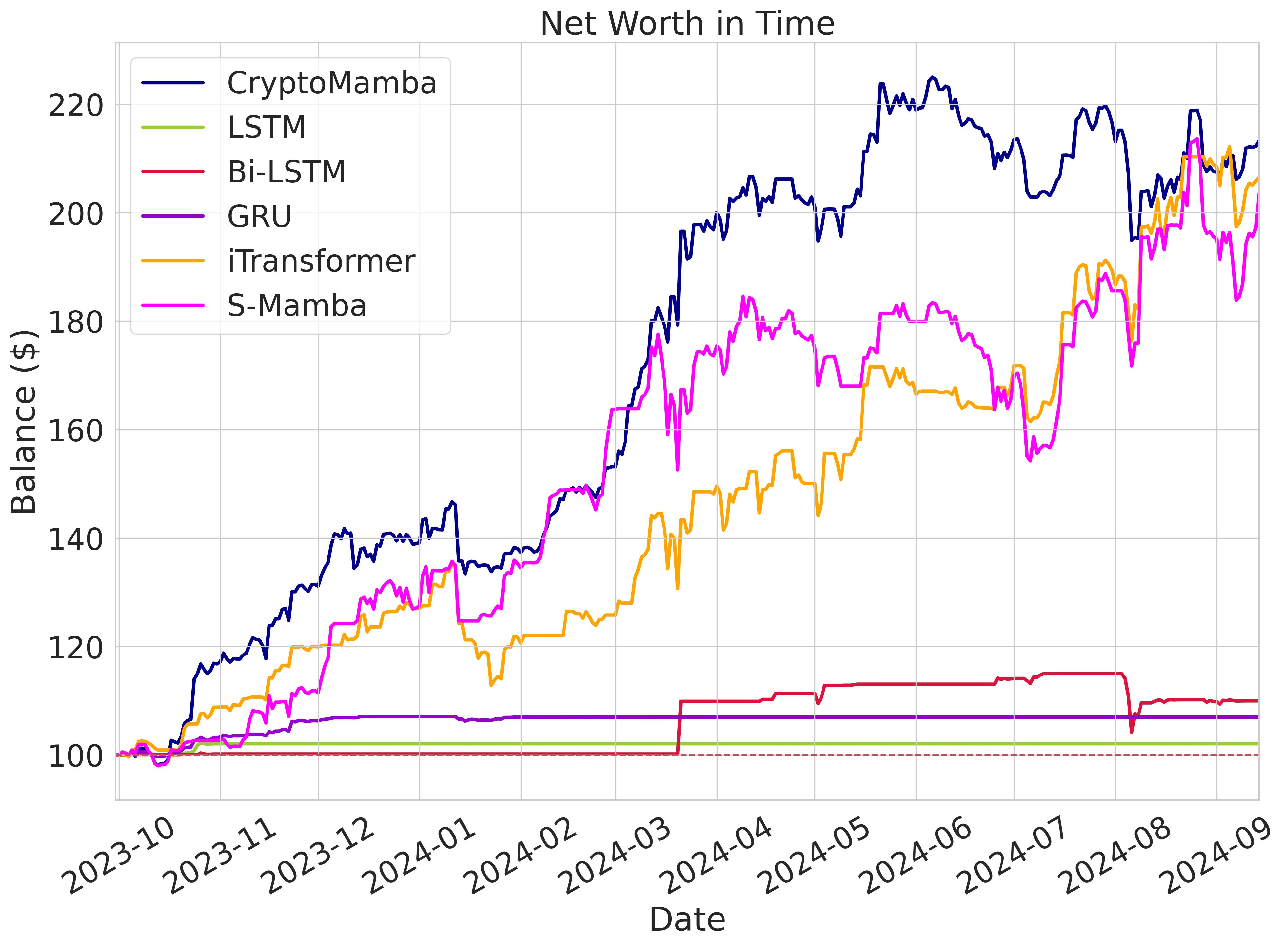}
        }
        \subfloat[Extended Smart - with volume]{
            \includegraphics[width=0.32\textwidth]{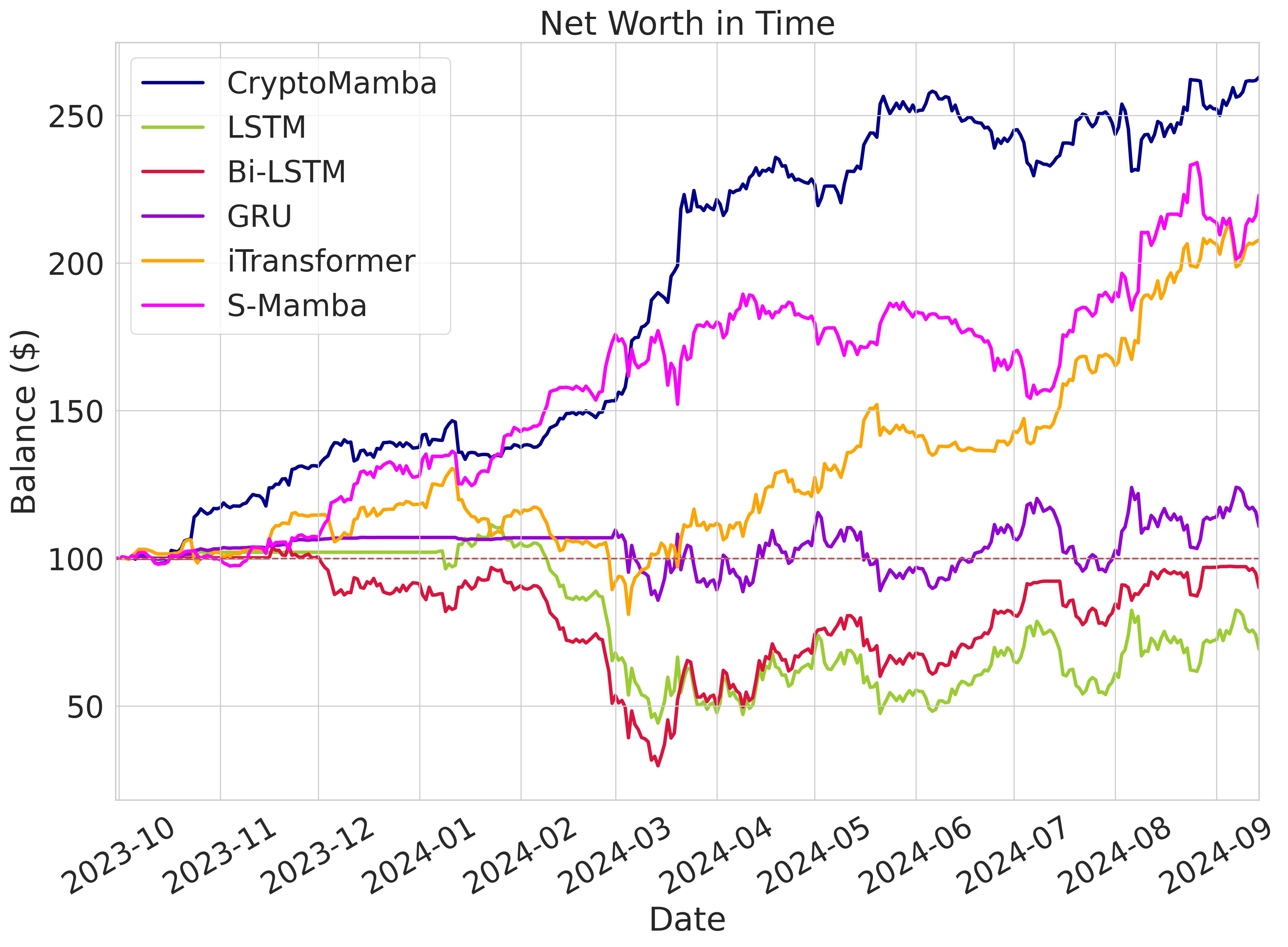}
        }
        \caption{Net worth over the test period using three trading strategies, Vanilla, Smart, and Extended Smart, under both volume-exclusive (top row) and volume-inclusive (bottom row) setups. \method{} consistently achieves the highest final balance across all configurations.}
	\label{fig:networth}
\end{figure*}

The trading results during the test period are presented in Figure \ref{fig:networth}, with the validation and test period final net worth and MDD summarized in Table \ref{table:final_balance}. In the test period, \method{} consistently achieves the highest returns in all three trading strategies, with \method{}-v ending with \$246.58 in the Vanilla, \$213.20 in the Smart, and \$262.78 in the Extended Smart setup, outperforming all baselines. It also consistently maintains a low MDD, especially in Extended Smart (11.09\%), indicating robust risk management. This highlights \method{}’s ability to translate accurate predictions into tangible financial gains while adapting well to both market rallies and downturns.

Traditional baselines like LSTM and Bi-LSTM perform relatively well in stable conditions such as the validation period but fail to generalize in the test period, where high volatility and sharp reversals occur. This can be attributed to their lack of generalization and tendency to predict lower prices in the test period (Figures \ref{fig:no_vol} and \ref{fig:vol}). S-Mamba performs better in volatile settings, benefiting from its long-range modeling capacity, yet underperforms in stable intervals. The Extended Smart strategy further emphasizes the weaknesses of simpler baselines, with many models exhibiting high drawdown and final balances below the initial investment (e.g., LSTM: \$62.04, MDD: 71.46\%). In contrast, \method{} maintains strong performance across all setups, balancing returns and drawdowns effectively.

\begin{table}[htbp]
    \centering
    \caption{\centering The final balance and maximum drawdown (MDD) after one year of trading with \$100 in the validation and test period. Final balances that are lower than the initial investment are in \textit{Italic}.}
    \label{table:final_balance}
    \renewcommand{\arraystretch}{1.3} % Adjust row height
    \resizebox{ \textwidth}{!}{%
    \begin{tabular}{lcccccccccccccccccc}
    \toprule
    
    && \multicolumn{8}{c}{\textbf{Validation}} && \multicolumn{8}{c}{\textbf{Test}} \\
    \cmidrule{3-10} \cmidrule{12-19}
    
    && \multicolumn{2}{c}{\textbf{Vanilla}} && \multicolumn{2}{c}{\textbf{Smart}} && \multicolumn{2}{c}{\textbf{Extended Smart}} && \multicolumn{2}{c}{\textbf{Vanilla}} && \multicolumn{2}{c}{\textbf{Smart}} && \multicolumn{2}{c}{\textbf{Extended Smart}} \\
    \cmidrule{3-4} \cmidrule{6-7} \cmidrule{9-10} \cmidrule{12-13} \cmidrule{15-16} \cmidrule{18-19}
    \textbf{Model} && {FB(\$)} & {MDD(\%)} && {FB(\$)} & {MDD(\%)} && {FB(\$)} & {MDD(\%)} && {FB(\$)} & {MDD(\%)} && {FB(\$)} & {MDD(\%)} && {FB(\$)} & {MDD(\%)}\\ 
    % \midrule
    \toprule
    LSTM && 133.10 & 25.38 && 128.93 & 24.09 && 128.93 & 24.09 && 100.00 & 0.00 && 103.66 & 0.25 && \textit{62.04} & 71.46\\
    LSTM-v && 136.93 & 25.82 && \textbf{149.95} & 25.82 && \textbf{149.95} & 25.82 && 100.00 & 0.00 && 102.06 & 0.20 && \textit{70.37} & 60.3\\
    Bi-LSTM && 134.78 & 25.82 && 141.36 & 25.82 && 141.36 & 25.82 && 100.00 & 0.00 && 100.00 & 0.00 && \textit{67.97} & 61.63\\
    Bi-LSTM-v && \textbf{156.98} & 25.38 && 134.41 & 25.01 && 136.37 & 25.01 && \textit{98.87} & 12.59 && 109.96 & 9.39 && \textit{91.21} & 71.46\\
    GRU && 115.12 & 19.80 && 109.06 & 19.32 && 111.59 & 19.32 && 100.00 & 0.00 && 112.64 & 3.79 && 117.64 & 21.78\\
    GRU-v && 153.57 & 18.70 && 125.04 & 22.31 && 128.07 & 22.31 && 100.00 & 0.00 && 106.96 & 0.88 && 111.96 & 22.85\\
    iTransformer && 100.80 & 28.21 && 114.30 & 20.25 && 103.50 & 24.06 && 151.91 & 14.01 && 151.58 & 13.00 && 128.53 & 28.27\\
    iTransformer-v && 100.19 & 24.70 && 116.95 & 20.30 && 115.50 & 21.75 && 241.50 & 17.94 && 206.38 & 16.57  && 207.67 & 37.86\\
    S-Mamba && \textit{75.94} & 31.32 && \textit{86.77} & 21.27 && \textit{82.95} & 26.59 && 146.08 & 33.01 && 173.00 & 24.24 && 155.66 & 28.27\\
    S-Mamba-v && \textit{77.25} & 35.63 && \textit{91.05} & 21.42 && \textit{78.03} & 34.29 && 182.63 & 27.75 && 203.17 & 16.45 && 222.50 & 18.61\\
    \midrule
    CryptoMamba && 113.24 & 24.11 && 106.80 & 22.82 && 102.99 & 22.86 && 175.05 & 23.33 && 184.56 & 16.01 && 204.04 & 17.95\\
    CryptoMamba-v && 124.09 & 25.38 && 127.12 & 23.40 && 127.47 & 23.40 && \textbf{246.58} & 23.48 && \textbf{213.20} & 13.37 && \textbf{262.78} & 11.09\\
    % \midrule
    % Buy and Hold && 124.09 & - && 127.12 & - && \textbf{246.58} & - && \textbf{213.20} & -\\
    \bottomrule
    \end{tabular}
        }
\end{table}

\method{} exhibits superior generalization across all scenarios. Unlike the baselines, \method{} performs consistently well in both steady and volatile intervals. Its robust design allows it to adapt dynamically to market trends, achieving consistent profitability in all setups and outperforming state-of-the-art baselines in both the validation and test periods.

These results emphasize the practical utility of \method{} for real-world financial applications. By combining accurate predictions with trading strategies, \method{} demonstrates its reliability and adaptability across diverse market conditions for financial forecasting.

%% file: sections/7_conclusion.tex
\section{Conclusion and Future Work}

In this work, we introduced \method{}, a novel Mamba-based architecture tailored for financial time-series forecasting, and demonstrated its effectiveness in predicting Bitcoin prices. Leveraging state space models (SSMs) with Mamba enhancements, \method{} efficiently captures long-range dependencies and outperforms traditional models such as LSTM, Bi-LSTM, GRU, iTransformer, and S-Mamba in both predictive accuracy and financial performance. Our results show that \method{}, especially in its volume-inclusive variant, achieves the highest returns in real-world trading scenarios, highlighting the importance of incorporating volume data for enhanced market insights.

The trading simulations using Vanilla, Smart, and Extended Smart algorithms emphasize the practical utility of \method{} in real-world trading. By employing trading algorithms as evaluation tools, we were able to effectively compare prediction models in terms of their financial impact, moving beyond regression metrics alone. These results highlight the versatility of \method{} in adapting to different market conditions and underline its ability to achieve profitability even under highly volatile scenarios.
\method{} design not only proves its capability in modeling cryptocurrency markets but also showcases its potential for forecasting other time series data, such as stocks or commodities, where long-range dependencies are critical.

Future work could explore extending \method{} to other financial assets and testing its adaptability to multi-asset portfolios. Additionally, refining trading algorithms to better leverage prediction intervals and incorporating external factors, such as sentiment analysis or macroeconomic indicators, may further improve performance. While \method{} demonstrates strong generalization and profitability, further research into integrating risk management strategies within the architecture itself could provide even greater robustness in highly volatile markets.

%% file: sections/8_appendix.tex
\section*{Appendix}

\section{Metrics}\label{sec:metrics}

\paragraph{Regression Metrics.}
For evaluating the forecasting accuracy of our models, we use three standard metrics: Root Mean Squared Error (RMSE), Mean Absolute Percentage Error (MAPE), and Mean Absolute Error (MAE). The formulas for these metrics are presented in Table \ref{table:loss}, where \( y_i \) represents the actual value, \( \hat{y}_i \) is the predicted value, and \( n \) is the total number of predictions.

\begin{table}[h]
\centering
\caption{Definitions of the evaluation metrics}
\label{table:loss}
\renewcommand{\arraystretch}{1.4} 
\resizebox{0.32\textwidth}{!}{
\begin{tabular}{cc}
\toprule
\textbf{Metric} & \textbf{Formula} \\
\midrule
RMSE & $\sqrt{\frac{1}{n} \sum_{i=1}^{n} (y_i - \hat{y}_i)^2}$\\
MAPE & $\frac{100}{n} \sum_{i=1}^{n} \left| \frac{y_i - \hat{y}_i}{y_i} \right|$\\
MAE & $\frac{1}{n} \sum_{i=1}^{n} |y_i - \hat{y}_i|$\\
\bottomrule
\end{tabular}
}
\end{table}

RMSE penalizes larger errors more heavily due to the squaring of deviations, making it particularly useful in scenarios where large forecasting errors are costly. MAPE expresses error as a percentage, offering scale-independence and making it particularly valuable when comparing across different ranges, as Bitcoin prices can vary widely over time due to high volatility, and different time intervals may exhibit vastly different price ranges. However, it can inflate errors when actual values are small. MAE measures the average absolute error, treating all deviations equally, providing a robust and balanced measure of accuracy without overemphasizing outliers.

Lower values across all three metrics indicate better performance, with RMSE and MAE measuring absolute error and MAPE capturing relative error. Together, these metrics provide a well-rounded evaluation of forecasting accuracy from multiple perspectives.

\paragraph{Maximum Drawdown (MDD).}
Maximum Drawdown (MDD) quantifies the largest peak-to-trough decline in a portfolio's value over a given time period. It measures the worst possible loss an investor could have faced if buying at a local maximum and selling at the subsequent minimum. MDD is defined as:

\begin{equation}
\text{MDD} = \max_{t \in [1, T]} \left( \frac{\max_{i \in [1, t]} P_i - P_t}{\max_{i \in [1, t]} P_i} \right)
\end{equation}

where:
\begin{itemize}
    \item \( P_t \) is the portfolio value at time step \( t \),
    \item \( T \) is the total number of time steps,
    \item \( \max_{i \in [1, t]} P_i \) is the historical peak value up to time \( t \).
\end{itemize}

This metric expresses the largest percentage drop from a historical peak, helping assess downside risk and the volatility of a trading strategy. Lower MDD values indicate better risk control.

\section{Extended Smart Trading Algorithm}\label{sec:extended_smart}

The \textbf{Extended Smart Trading Algorithm} builds upon the Smart strategy by incorporating short-selling capabilities, enabling the model to profit even during market downturns. Like the Smart algorithm, it defines a risk-based interval \([y_{\min}, y_{\max}]\) around the predicted price \(y\) and adjusts the trading action based on the current price \(x\) relative to this interval. However, in scenarios where the predicted price is lower than today's price, the algorithm can go short (i.e., sell more than the currently held bitcoin). To prevent excessive risk exposure, we introduce a \textit{maximum short position} limit, restricting the amount of negative bitcoin the strategy can hold to 0.002 BTC. This constraint ensures controlled risk while enabling more aggressive trading decisions that can outperform traditional long-only strategies in volatile or bearish markets. The detailed steps of this algorithm are provided in Algorithm~\ref{alg:extended_smart_trading}.

\begin{algorithm}[H]
\label{alg:three}
\caption{Extended Smart Trading Algorithm}
\label{alg:extended_smart_trading}
\begin{algorithmic}[1]
\REQUIRE Prediction, Today's Price, risk, maximum negative bitcoin
\STATE $x \gets \text{Today's price}$
\STATE $y \gets \text{Prediction}$
\STATE $y_{\max} \gets \left(1 + \frac{\text{risk}}{100}\right)y$
\STATE $y_{\min} \gets \left(1 - \frac{\text{risk}}{100}\right)y$
\IF{$x \geq y$}
    \IF{$x \geq y_{\max}$}
        \STATE Sell as much as shares to reach the maximum short position
    \ELSE
        \IF{you have positive shares}
            \STATE Sell $\frac{x - y}{y_{\max} - y}$ of your shares
        \ENDIF
        % \ELSE
        %     \STATE {Do nothing}
    \ENDIF
\ELSE
    \IF{$x \leq y_{\min}$}
        \STATE Buy with all of your money
    \ELSE
        \STATE {Buy with $\frac{y - x}{y - y_{\min}}$ of your money}
    \ENDIF
\ENDIF
\end{algorithmic}
\end{algorithm}

% \begin{table}[h!]
%     \centering
%     \caption{\centering The final balance and maximum drawdown (MDD) after one year of trading with \$100 in the validation and test period. Final balances that are lower than the initial investment are in \textit{Italic} font.}
%     \label{table:final_balance_esmart}
%     \renewcommand{\arraystretch}{1.3} % Adjust row height
%     % \setlength{\tabcolsep}{10pt} % Adjust column spacing
%     \resizebox{0.5\textwidth}{!}{%
%     \begin{tabular}{lcccccc}
%     \toprule
    
%     && \multicolumn{2}{c}{\textbf{Validation}} && \multicolumn{2}{c}{\textbf{Test}} \\
%     \cmidrule{3-4} \cmidrule{6-7}
    
%     \textbf{Model} && {FB(\$)} & {MDD(\%)} && {FB(\$)} & {MDD(\%)}\\ 
%     \toprule
%     LSTM && 128.93 & 24.09 && \textit{62.04} & 71.46\\
%     LSTM-v && \textbf{149.95} & 25.82 && \textit{70.37} & 60.3\\
%     Bi-LSTM && 141.36 & 25.82 && \textit{67.97} & 61.63\\
%     Bi-LSTM-v && 136.37 & 25.01 && \textit{91.21} & 71.46\\
%     GRU && 111.59 & 19.32 && 117.64 & 21.78 \\
%     GRU-v && 128.07 & 22.31 && 111.96 & 22.85 \\
%     iTransformer && 103.50 & 24.06 && 128.53 & 28.27\\
%     iTransformer-v && 115.50 & 21.75 && 207.67 & 37.86 \\
%     S-Mamba && \textit{82.95} & 26.59 && 155.66 & 28.27\\
%     S-Mamba-v && \textit{78.03} & 34.29 && 222.50 & 18.61\\
%     \midrule
%     CryptoMamba && 102.99 & 22.86 && 204.04 & 17.95\\
%     CryptoMamba-v && 127.47 & 23.4 && \textbf{262.78} & 11.09 \\
%     % \midrule
%     % Buy and Hold && 124.09 & - && 127.12 & - && \textbf{246.58} & - && \textbf{213.20} & -\\
%     \bottomrule
%     \end{tabular}
%         }
% \end{table}